%% file: chapter22-Thibeau-Sutre_Collin-interpretability.tex
\documentclass[a4paper, twoside, 12pt]{article}

\input{template}

\begin{document}


\newcommand{\runningauthor}{Thibeau-Sutre \textit{et al.}} 

\newcommand{\runningheadtitle}{Interpretability}

\newcommand{\chapternumber}{22}

\newcommand{\emailaddress}{elina.ts@free.fr}

\title{Interpretability of Machine Learning Methods Applied to Neuroimaging} 

\author[1*]{Elina Thibeau-Sutre}
\author[1]{Sasha Collin}
\author[1]{Ninon Burgos}
\author[1]{Olivier Colliot}

\affil[1]{Sorbonne Universit\'e, Institut du Cerveau - Paris Brain Institute - ICM, CNRS, Inria, Inserm, AP-HP, H\^opital de la Piti\'e-Salp\^etri\`ere, F-75013, Paris, France}

\affil[*]{Corresponding author: e-mail address: \href{mailto:\emailaddress}{\emailaddress}}

\maketitle

\afterpage{\aftergroup\restoregeometry}
\pagestyle{otherpages}

\begin{abstract}
Deep learning methods have become very popular for the processing of natural images, and were then successfully adapted to the neuroimaging field. As these methods are non-transparent, interpretability methods are needed to validate them and ensure their reliability. Indeed, it has been shown that deep learning models may obtain high performance even when using irrelevant features, by exploiting biases in the training set. Such undesirable situations can potentially be detected by using interpretability methods. Recently, many methods have been proposed to interpret neural networks. However, this domain is not mature yet.
Machine learning users face two major issues when aiming to interpret their models: which method to choose, and how to assess its reliability? Here, we aim at providing answers to these questions by presenting the most common interpretability methods and metrics developed to assess their reliability, as well as their applications and benchmarks in the neuroimaging context. Note that this is not an exhaustive survey: we aimed to focus on the studies which we found to be the most representative and relevant.

\end{abstract}

\begin{keywords}
interpretability, saliency, machine learning, deep learning, neuroimaging, brain disorders
\end{keywords}

\section{Introduction}
\label{sec:intro}

\subsection{Need for interpretability} 

Many metrics have been developed to evaluate the performance of machine learning (ML) systems. In the case of supervised systems, these metrics compare the output of the algorithm to a ground truth, in order to evaluate its ability to reproduce a label given by a physician. However, the users (patients and clinicians) may want more information before relying on such systems. On which features is the model relying to compute the results? Are these features close to the way a clinician thinks? If not, why? This questioning coming from the actors of the medical field is justified, as errors in real life may lead to dramatic consequences.
Trust into ML systems cannot be built only based on a set of metrics evaluating the performance of the system.
Indeed, various examples of machine learning systems taking correct decisions for the wrong reasons exist, e.g. ~\cite{ribeiroWhyShouldTrust2016,fongInterpretableExplanationsBlack2017,degraveAIRadiographicCOVID192021}. Thus, even though their performance is high, they may be unreliable and, for instance, not generalize well to slightly different data sets. One can try to prevent this issue by interpreting the model with an appropriate method whose output will highlight the reasons why a model took its decision. 

\begin{figure}[!tbh]
    \centering
    \includegraphics[width=0.6\textwidth]{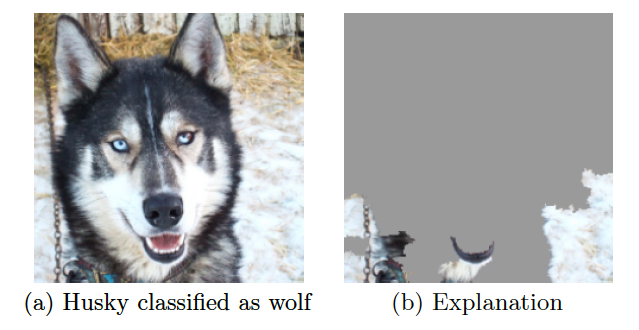}
    \caption[Example of an interpretability method highlighting why a network took the wrong decision.]{Example of an interpretability method highlighting why a network took the wrong decision. The explained classifier was trained on the binary task ``Husky'' vs ``Wolf''. The pixels used by the model are actually in the background and highlight the snow. \\
    Adapted from \citep{ribeiroWhyShouldTrust2016}. Permission to reuse was kindly granted by the authors.}
    \label{fig: ribeiro_husky_snow}
\end{figure}

In \cite{ribeiroWhyShouldTrust2016}, the authors show a now classical case of a system that correctly classifies images for wrong reasons. They purposely designed a biased data set in which wolves always are in a snowy environment whereas huskies are not. Then, they trained a classifier to differentiate wolves from huskies: this classifier had a good accuracy, but classified as wolves huskies with a snowy background, and as huskies wolves that were not in the snow. Using an interpretability method, they further highlighted that the classifier was looking at the background and not at the animal (see Figure~\ref{fig: ribeiro_husky_snow}).

Another study \cite{fongInterpretableExplanationsBlack2017} detected a bias in ImageNet (a widely used data set of natural images) as the interpretation of images with the label ``chocolate sauce'' highlighted the importance of the spoon. Indeed, ImageNet ``chocolate sauce'' images often contained spoons, leading to a spurious correlation. There are also examples of similar problems in medical applications. For instance, a recent paper \cite{degraveAIRadiographicCOVID192021} showed with interpretability methods that some deep learning systems detecting COVID-19 from chest radiographs actually relied on confounding factors rather than on the actual pathological features. Indeed, their model focused on other regions than the lungs to evaluate the COVID-19 status (edges, diaphragm and cardiac silhouette).  Of note, their model was trained on public data sets which were used by many studies. 

\subsection{How to interpret models} 

According to \cite{liptonMythosModelInterpretability2018}, model interpretability can be broken down into two categories: transparency and post-hoc explanations. 

A model can be considered as transparent when it (or all parts of it) can be fully understood as such, or when the learning process is understandable. A natural and common candidate that fits, at first sight, these criteria is the linear regression algorithm, where coefficients are usually seen as the individual contributions of the input features. Another candidate is the decision tree approach where model predictions can be broken down into a series of understandable operations. One can reasonably consider these models as transparent: one can easily identify the features that were used to take the decision. However, one may need to be cautious not to push too far the medical interpretation. Indeed, the fact that a feature has not been used by the model does not mean that it is not associated with the target. It just means that the model did not need it to increase its performance. For instance, a classifier aiming at diagnosing Alzheimer's disease may need only a set of regions (for instance from the medial temporal lobe of the brain) to achieve an optimal performance. This does not mean that other brain regions are not affected by the disease, just that they were not used by the model to take its decision. This is the case for example for sparse models like LASSO, but also standard multiple linear regressions. Moreover, features given as input to transparent models are often highly-engineered, and choices made before the training step (preprocessing, feature selection) may also hurt the transparency of the whole framework. Nevertheless, in spite of these caveats, such models can reasonably be considered transparent, in particular when compared to deep neural networks which are intrinsically black boxes.


The second category of interpretability methods, post-hoc interpretations, allows dealing with non-transparent models. Xie et al.~\cite{xieExplainableDeepLearning2020} proposed a taxonomy in three categories: \textit{visualization} methods consist in extracting an attribution map of the same size as the input whose intensities allow knowing where the algorithm focused its attention, \textit{distillation} approaches consist in reproducing the behavior of a black-box model with a transparent one, and \textit{intrinsic} strategies include interpretability components within the framework, which are trained along with the main task (for example, a classification). In the present work, we focus on this second category of methods (post-hoc), and proposed a new taxonomy including other methods of interpretation (see Figure~\ref{fig:taxonomy}). Post-hoc interpretability is the category the most used nowadays, as it allows interpreting deep learning methods that became the state-of-the-art for many tasks in neuroimaging, as in other application fields.

\subsection{Chapter content and outline}

This chapter focuses on methods developed to interpret non-transparent machine learning systems, mainly deep learning systems,
computing classification or regression tasks from high-dimensional inputs. The interpretability of other frameworks (in particular generative models such as variational autoencoders or generative adversarial networks) is not covered as there are not enough studies addressing them. It may be because high-dimensional outputs (such as images) are easier to interpret ``as such'', whereas small dimensional outputs (such as scalars) are less transparent.

Most interpretability methods presented in this chapter produce an attribution map: an array with the same dimensions as that of the input (up to a resizing), that can be overlaid on top of the input in order to exhibit an explanation of the model prediction. In the literature, many different terms may coexist to name this output such as saliency map, interpretation map or heatmap. To avoid misunderstandings, in the following, we will only use the term ``attribution map''.

The chapter is organized as follows. Section~\ref{sec:section2} presents the most commonly used interpretability methods proposed for computer vision, independently of medical applications. It also describes metrics developed to evaluate the reliability of interpretability methods. Then, section~\ref{sec:section3} details their application to neuroimaging. Finally, section~\ref{sec:limitations} discusses current limitations of interpretability methods, presents benchmarks conducted in the neuroimaging field and gives some advice to the readers who would like to interpret their own models.

Mathematical notations and abbreviations used during this chapter are summarized in Table~\ref{tab:notations} and Table~\ref{tab:abbreviations}. A short reminder on neural network training procedure and a brief description of the diseases mentioned in the present chapter are provided in Appendices~\ref{appendix:network} and~\ref{appendix:diseases}.

\begin{table}
    \caption{Mathematical notations}
    \label{tab:notations}
    \noindent\hrulefill
    \begin{itemize}
    \item $X_0$ is the input tensor given to the network, and $X$ refers to any input, sampled from the set $\mathcal{X}$.
    \item $y$ is a vector of target classes corresponding to the input.
    \item $f$ is a network of $L$ layers. The first layer is the closest to the input, the last layer is the closest to the output. A layer is a function.
    \item $g$ is a transparent function which aims at reproducing the behaviour of $f$.
    \item $w$ and $b$ are the weights and the bias associated to a linear function (for example in a fully-connected layer).
    \item $u$ and $v$ are locations (set of coordinates) corresponding to a node in a feature map. They belong respectively to the set $\mathcal{U}$ and $\mathcal{V}$.
    \item $A^{(l)}_k(u)$ is the value of the feature map computed by layer $l$, of $K$ channels at channel $k$, at position $u$.
    \item $R^{(l)}_k(u)$ is the value of a property back-propagated through the $l+1$, of $K$ channels at channel $k$, at position $u$. $R^{(l)}$ and $A^{(l)}$ have the same number of channels.
    \item $o_c$ is the output node of interest (in a classification framework, it corresponds to the node of the class $c$).
    \item $S_c$ is an attribution map corresponding to the output node $o_c$.
    \item $m$ is a mask of perturbations. It can be applied to $X$ to compute its perturbed version $X^m$.
    \item $\Phi$ is a function producing a perturbed version of an input $X$.
    \item $\Gamma_c$ is the function computing the attribution map $S_c$ from the black-box function $f$ and an input $X_0$.
    \end{itemize}
    \noindent\hrulefill
\end{table}

\begin{table}
    \caption{Abbreviations}
    \label{tab:abbreviations}
    \noindent\hrulefill
    \begin{itemize}
        \item \textbf{CAM} Class activation maps
        \item \textbf{CNN} Convolutional neural network
        \item \textbf{CT} Computed tomography
        \item \textbf{Grad-CAM} Gradient-weighted class activation mapping
        \item \textbf{LIME} Local interpretable model-agnostic explanations
        \item \textbf{LRP} Layer-wise relevance
        \item \textbf{MRI} Magnetic resonance imaging
        \item \textbf{SHAP} SHapley Additive exPlanations
        \item \textbf{T1w} T1-weighted [Magnetic Resonance Imaging]
    \end{itemize}
    \noindent\hrulefill
\end{table}

\section{Interpretability methods} 
\label{sec:section2}

This section presents the main interpretability methods proposed in the domain of computer vision. We restrict ourselves to the methods that have been applied to the neuroimaging domain (the applications themselves being presented in Section~\ref{sec:section3}).
The outline of this section is largely inspired from the one proposed by Xie et al.~\cite{xieExplainableDeepLearning2020}:
\begin{enumerate}
    \item \textbf{weight visualization} consists in directly visualizing weights learned by the model, which is natural for linear models but quite less informative for deep learning networks,
    \item \textbf{feature map visualization} consists in displaying intermediate results produced by a deep learning network to better understand its operation principle,
    \item \textbf{back-propagation methods} back-propagate a signal through the machine learning system from the output node of interest $o_c$ to the level of the input to produce an attribution map,
    \item \textbf{perturbation methods} locally perturb the input and evaluate the difference in performance between using the original input and the perturbed version to infer which parts of the input are relevant for the machine learning system,
    \item \textbf{distillation} approximates the behavior of a black-box model with a more transparent one, and then draw conclusions from this new model,
    \item \textbf{intrinsic} includes the only methods of this chapter that are not post-hoc explanations: in this case, interpretability is obtained thanks to components of the framework that are trained at the same time as the model. 
\end{enumerate}
%
%
Finally, for the methods producing an attribution map, a section is dedicated to the metrics used to evaluate different properties (for example reliability or human-intelligibility) of the maps.

We caution readers that this taxonomy is not perfect: some methods may belong to several categories (for example LIME and SHAP could belong either to perturbation or distillation methods). Moreover, interpretability is still an active research field, then some categories may (dis)appear or be fused in the future.

The interpretability methods were (most of the time) originally proposed in the context of a classification task. In this case, the network outputs an array of size $C$, corresponding to the number of different labels existing in the data set, and the goal is to know how the output node corresponding to a particular class $c$ interacts with the input or with other parts of the network. However, these techniques can be extended to other tasks: for example for a regression task, we will just have to consider the output node containing the continuous variable learned by the network. Moreover, some methods do not depend on the nature of the algorithm (e.g. standard-perturbation or LIME) and can be applied to any machine learning algorithm.

\tikzstyle{root}=[rectangle, draw=black, rounded corners, fill=lightgray, drop shadow, text centered, anchor=north, text=black, text width=5cm, font=\small]
\tikzstyle{category}=[rectangle, draw=black, rounded corners, fill=lightgray, drop shadow, text centered, anchor=north, text=black, text width=3.5cm, font=\footnotesize]
\tikzstyle{description}=[rectangle, draw=black, rounded corners, fill=white, drop shadow, anchor=north, text=black, text width=3.5cm, font=\scriptsize]
\tikzstyle{myarrow}=[stealth-, thick]

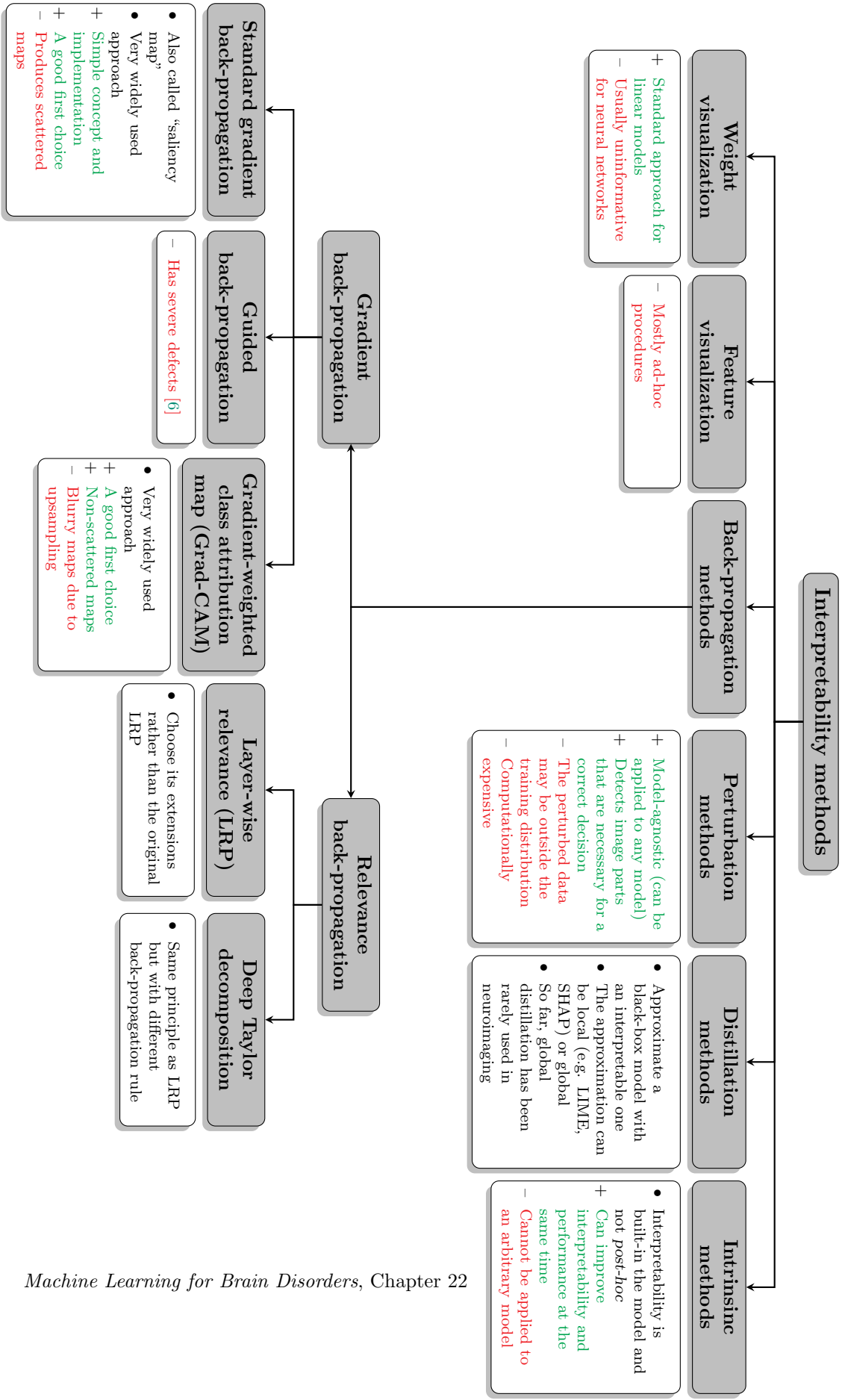
\begin{sidewaysfigure}
\centering
\begin{tikzpicture}[node distance=2cm]
    \node (Root) [root]
        {
            \textbf{Interpretability methods}
        };

        \node (Weight) [category, below left=1cm and 5.5cm of Root]
            {
                \textbf{Weight \\visualization}
            };
            \node (WeightDescription) [description, below=0.2cm of Weight]
            {
            \begin{itemize}[leftmargin=*]
                \item[+] \textcolor{Green}{Standard approach for linear models}
                \item[--] \textcolor{Red}{Usually uninformative for neural networks}
            \end{itemize}
            };

        \node (Feature) [category, below left=1cm and 1.5cm of Root]
            {
                \textbf{Feature \\visualization}
            };
            \node (FeatureDescription) [description, below=0.2cm of Feature]
            {
            \begin{itemize}[leftmargin=*]
                \item[--] \textcolor{Red}{Mostly ad-hoc procedures}
            \end{itemize}    
            };

        \node (Backprop) [category, below left=1cm and -2.5cm of Root]
            {
                \textbf{Back-propagation \\methods}
            };

            \node (Gradient) [category, below left=5.5cm and 1cm of Backprop]
                {
                    \textbf{Gradient \\back-propagation}
                };
                \node (Standard) [category, below left=1cm and 0.25cm of Gradient]
                    {
                        \textbf{Standard gradient back-propagation}
                    };
                \node (StandardDescription) [description, below=0.2cm of Standard]
                    {
                    \begin{itemize}[leftmargin=*]
                        \item Also called ``saliency map''
                        \item Very widely used approach
                        \item[+] \textcolor{Green}{Simple concept and implementation}
                        \item[+] \textcolor{Green}{A good first choice}
                        \item[--] \textcolor{Red}{Produces scattered maps}
                    \end{itemize} 
                    };
                \node (Guided) [category, below=1cm of Gradient]
                    {
                        \textbf{Guided \\back-propagation}
                    };
                \node (GuidedDescription) [description, below=0.2cm of Guided]
                    {
                    \begin{itemize}[leftmargin=*]
                        \item[--] \textcolor{Red}{Has severe defects \cite{adebayoSanityChecksSaliency2018}}
                    \end{itemize}                      };
                \node (GradCAM) [category, below right=1cm and 0.25cm of Gradient]
                    {
                        \textbf{Gradient-weighted class attribution map (Grad-CAM)}
                    };
                \node (GradCAMDescription) [description, below=0.2cm of GradCAM]
                    {
                    \begin{itemize}[leftmargin=*]
                        \item Very widely used approach
                        \item[+] \textcolor{Green}{A good first choice}
                        \item[+] \textcolor{Green}{Non-scattered maps}
                        \item[--] \textcolor{Red}{Blurry maps due to upsampling}
                    \end{itemize}                     
                    };
            \node (Relevance) [category, below right=5.5cm and 1.5cm of Backprop]
                {
                    \textbf{Relevance \\back-propagation}
                };
                \node (LRP) [category, below left=1cm and -1.75cm of Relevance]
                    {
                        \textbf{Layer-wise \\relevance (LRP)}
                    };
                \node (LRPDescription) [description, below=0.2cm of LRP]
                    {
                    \begin{itemize}[leftmargin=*]
                        \item Choose its extensions rather than the original LRP
                    \end{itemize}                      };
                \node (Taylor) [category, below right=1cm and -1.75cm of Relevance]
                    {
                        \textbf{Deep Taylor decomposition}
                    };
                 \node (TaylorDescription) [description, below=0.2cm of Taylor]
                    {
                    \begin{itemize}[leftmargin=*]
                        \item Same principle as LRP but with different back-propagation rule
                    \end{itemize}
                    };

        \node (Perturbation) [category, below right=1cm and -2.5cm of Root]
            {
                \textbf{Perturbation \\methods}
            };
            \node (PerturbationDescription) [description, below=0.2cm of Perturbation]
            {
            \begin{itemize}[leftmargin=*]
                \item[+] \textcolor{Green}{Model-agnostic (can be applied to any model)}
                \item[+] \textcolor{Green}{Detects image parts that are necessary for a correct decision}
                \item[--] \textcolor{Red}{The perturbed data may be outside the training distribution}
                \item[--] \textcolor{Red}{Computationally expensive}
            \end{itemize}
            };

        \node (Distillation) [category, below right=1cm and 1.5cm of Root]
            {
                \textbf{Distillation \\methods}
            };
            \node (DistillationDescription) [description, below=0.2cm of Distillation]
            {
            \begin{itemize}[leftmargin=*]
                    \item Approximate a black-box model with an interpretable one
                    \item The approximation can be local (e.g. LIME, SHAP) or global
                    \item So far, global distillation has been rarely used in neuroimaging
            \end{itemize}              
            };

        \node (Intrinsinc) [category, below right=1cm and 5.5cm of Root]
            {
                \textbf{Intrinsinc \\methods}
            };
            \node (IntrinsincDescription) [description, below=0.2cm of Intrinsinc]
            {
            \begin{itemize}[leftmargin=*]
                    \item Interpretability is built-in the model and not {\sl post-hoc}
                    \item[+] \textcolor{Green}{Can improve interpretability and performance at the same time}
                    \item[--] \textcolor{Red}{Cannot be applied to an arbitrary model}
            \end{itemize}  
            };

    \draw[myarrow] (Weight.north) -- ++(0,0.5) -| (Root.south);
    \draw[myarrow] (Feature.north) -- ++(0,0.5) -| (Root.south);
    \draw[myarrow] (Backprop.north) -- ++(0,0.5) -| (Root.south);
    \draw[myarrow] (Perturbation.north) -- ++(0,0.5) -| (Root.south);
    \draw[myarrow] (Distillation.north) -- ++(0,0.5) -| (Root.south);
    \draw[myarrow] (Intrinsinc.north) -- ++(0,0.5) -| (Root.south);
    \draw[myarrow] (Gradient.east) -| (Backprop.south);
    \draw[myarrow] (Standard.north) -- ++(0,0.5) -| (Gradient.south);
    \draw[myarrow] (Guided.north) -- ++(0,0.5) -| (Gradient.south);
    \draw[myarrow] (GradCAM.north) -- ++(0,0.5) -| (Gradient.south);
    \draw[myarrow] (Relevance.west) -| (Backprop.south);
    \draw[myarrow] (LRP.north) -- ++(0,0.5) -| (Relevance.south);
    \draw[myarrow] (Taylor.north) -- ++(0,0.5) -| (Relevance.south);

\end{tikzpicture}

\caption{Taxonomy of the main interpretability methods.}
\label{fig:taxonomy}

\end{sidewaysfigure}

\subsection{Weight visualization}

At first sight, one of can be tempted to directly visualize the weights learned by the algorithm. This method is really simple, as it does not require further processing. However, even though it can make sense for linear models, it is not very informative for most networks unless they are specially designed for this interpretation.

\begin{figure}[!tbh]
    \centering
    \includegraphics[width=\textwidth]{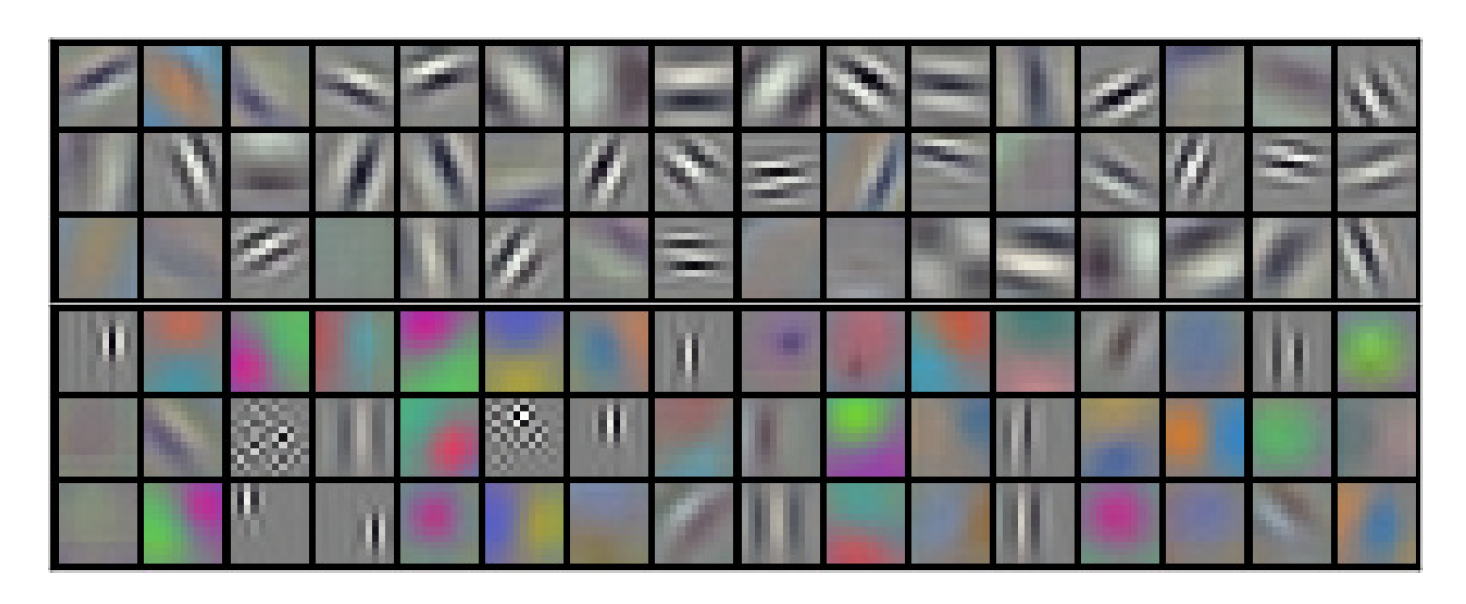}
    \caption[Convolutional kernels of learned by the first convolutional layer by AlexNet.]{96 convolutional kernels of size $3@11\times11$ learned by the first convolutional layer on the $3@224\times224$ input images by AlexNet. \\
    Adapted from \citep{krizhevskyImageNetClassificationDeep2012}. Permission to reuse was kindly granted by the authors.}
    \label{fig:alexnet_weights}
\end{figure}

This is the case for AlexNet \cite{krizhevskyImageNetClassificationDeep2012}, a convolutional neural network (CNN) trained on natural images (ImageNet). In this network the size of the kernels in the first layer is large enough ($11\times11$) to distinguish patterns of interest. Moreover, as the three channels in the first layer correspond to the three color channels of the images (red, green and blue), the values of the kernels can also be represented in terms of colors (this is not the case for hidden layers, in which the meaning of the channels is lost). The 96 kernels of the first layer were illustrated in the original article as in Figure~\ref{fig:alexnet_weights}. However, for hidden layers, this kind of interpretation may be misleading as non-linearity activation layers are added between the convolutions or fully-connected layers, this is why they only visualized the weights of the first layer.

\begin{figure}[!tbh]
    \centering
    \includegraphics[width=\textwidth]{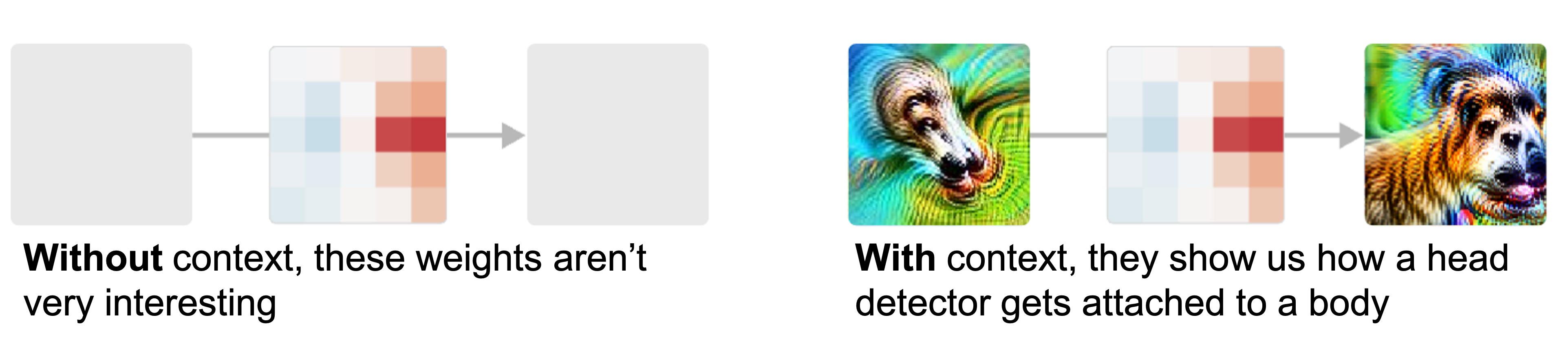}
    \caption[Weight visualization using feature maps context.]{The weights of small kernels in hidden layers (here $5\times5$) can be really difficult to interpret alone. Here some context allow better understanding how it modulates the interaction between concepts conveyed by the input and the output. \\
    Adapted from \citep{vossVisualizingWeights2021} (CC BY 4.0).}
    \label{fig:context_weights}
\end{figure}

To understand the weight visualization in hidden layers of a network, Voss et al.~\cite{vossVisualizingWeights2021} proposed to add some context to the input and the output channels. This way they enriched the weight visualization with feature visualization methods able to generate an image corresponding to the input node and the output node (see Figure~\ref{fig:context_weights}). However, the feature visualization methods used to bring some context can also be difficult to interpret themselves, then it only moves the interpretability problem from weights to features.

\subsection{Feature map visualization}

Feature maps are the results of intermediate computations done from the input and resulting in the output value. Then, it seems natural to visualize them, or link them to concepts to understand how the input is successively transformed into the output.

Methods described in this section aim at highlighting which concepts a feature map (or part of it) $A$ conveys.

\subsubsection{Direct interpretation}

The output of a convolution has the same shape as its input: a 2D image processed by a convolution will become another 2D image (the size may vary). Then, it is possible to directly visualize these feature maps and compare them to the input to understand the operations performed by the network. However, the number of filters of convolutional layers (often a hundred) makes the interpretation difficult as a high number of images must be interpreted for a single input.

Instead of directly visualizing the feature map $A$, it is possible to study the latent space including all the values of the samples of a data set at the level of the feature map $A$. Then, it is possible to study the deformations of the input by drawing trajectories between samples in this latent space, or more simply to look at the distribution of some label in a manifold learned from the latent space. In such a way, it is possible to better understand which patterns were detected, or at which layer in the network classes begin to be separated (in the classification case). There is often no theoretical framework to illustrate these techniques, then we referred to studies in the context of the medical application (see Section~\ref{sec:application_FM} for references).

\subsubsection{Input optimization}

\begin{figure}[!tbh]
    \centering
    \includegraphics[width=\textwidth]{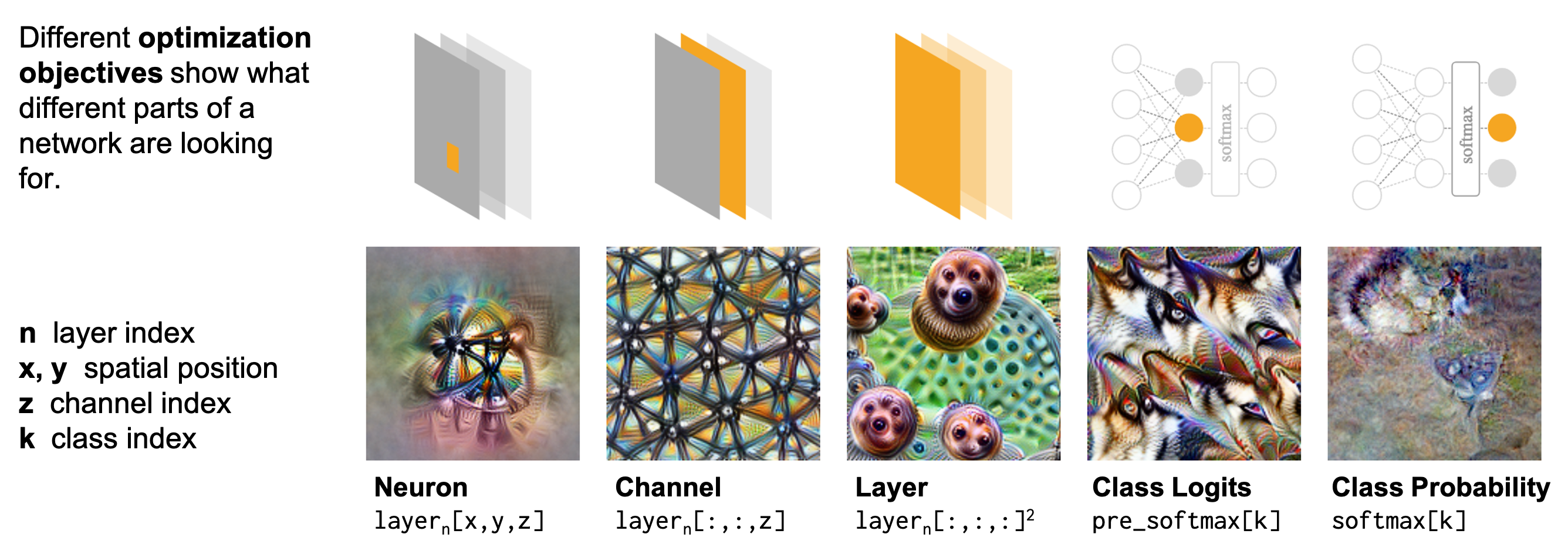}
    \caption[Optimization of the input for different levels of feature maps.]{Optimization of the input for different levels of feature maps. \\
    Adapted from \citep{olahFeatureVisualization2017a} (CC BY 4.0).}
    \label{fig:FM_parts}
\end{figure}

\begin{figure}[!tbh]
    \centering
    \includegraphics[width=\textwidth]{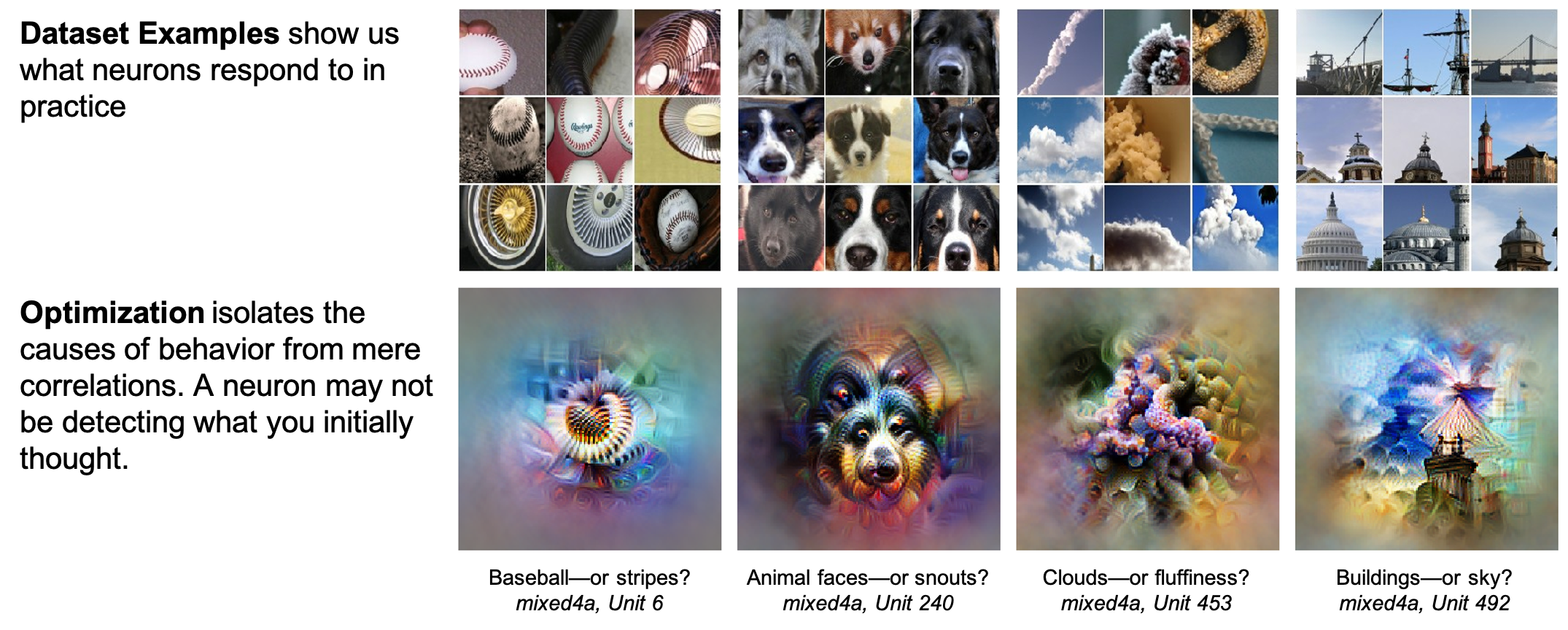}
    \caption[Association of input optimization with examples.]{Interpretation of a neuron of a feature map by optimizing the input associated with a bunch of training examples maximizing this neuron. \\
    Adapted from \citep{olahFeatureVisualization2017a} (CC BY 4.0).}
    \label{fig:FM_examples}
\end{figure}

Olah et al.~\cite{olahFeatureVisualization2017a} proposed to compute an input that maximizes the value of a feature map $A$ (see Figure~\ref{fig:FM_parts}). However, this technique leads to unrealistic images that may be themselves difficult to interpret, particularly for neuroimaging data. To have a better insight of the behavior of layers or filters, another simple technique illustrated by the same authors consists in isolating the inputs that led to the highest activation of $A$. The combination of both methods, displayed in Figure~\ref{fig:FM_examples}, allows a better understanding of the concepts conveyed by $A$ of a GoogleNet trained on natural images.

\subsection{Back-propagation methods}
\label{subsec:backprop}

The goal of these interpretability methods is to link the value of an output node of interest $o_c$ to the image $X_0$ given as input to a network. They do so by back-propagating a signal from $o_c$ to $X_0$: this process (backward pass) can be seen as the opposite operation than the one done when computing the output value from the input (forward pass).

Any property can be back-propagated as soon as its value at the level of a feature map $l-1$ can be computed from its value in the feature map $l$. In this section, the back-propagated properties are gradients or the relevance of a node $o_c$. 

\subsubsection{Gradient back-propagation}
 
During network training, gradients corresponding to each layer are computed according to the loss to update the weights. Then, we can see these gradients as the difference needed at the layer level to improve the final result: by adding this difference to the weights, the probability of the true class $y$ increases.

\begin{figure}
    \centering
    \includegraphics[width=0.8\textwidth]{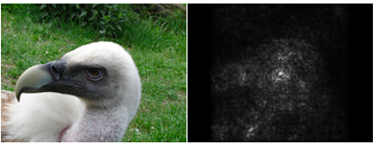}
    \caption[Attribution map of an image found with gradients back-propagation.]{Attribution map of an image found with gradients back-propagation.\\
    Adapted from \cite{simonyanDeepConvolutionalNetworks2013}. Permission to reuse was kindly granted by the authors.}
    \label{fig:simonyan_gradients}
\end{figure}

In the same way, the gradients can be computed at the image level to find how the input should vary to change the value of $o_c$ (see example on Figure~\ref{fig:simonyan_gradients}. This gradient computation was proposed by \cite{simonyanDeepConvolutionalNetworks2013}, in which the attribution map $S_c$ corresponding to the input image $X_0$ and the output node $o_c$ is computed according to the following equation:
\begin{equation}
    S_c = \frac{\partial{o_c}}{\partial{X}}\Bigr|_{\substack{X=X_0}}
\end{equation}

Due to its simplicity, this method is the most commonly used to interpret deep learning networks. Its attribution map is often called a ``saliency map'', however this term is also used in some articles to talk about any attribution map, and this is why we chose to avoid this term in this chapter.

This method was modified to derive many similar methods based on gradients computation described in the following paragraphs.

\paragraph{gradient$\odot$input}

This method is the point-wise product of the gradient map described at the beginning of the section and the input. Evaluated in \cite{shrikumarNotJustBlack2017a}, it was presented as an improvement of the gradients method, though the original paper does not give strong arguments on the nature of this improvement.

\paragraph{DeconvNet \& guided back-propagation}

The key difference between this procedure and the standard back-propagation method is the way the gradients are back-propagated through the ReLU layer.

The ReLU layer is a commonly used activation function that sets to 0 the negative input values, and does not affect positive input values. The derivative of this function in layer $l$ is the indicator function $\mathbb{1}_{A^{(l)}>0}$: it outputs 1 (resp. 0) where the feature maps computed during the forward pass were positive (resp. negative).

Springenberg et al.~\cite{springenbergStrivingSimplicityAll2014} proposed to back propagate the signal differently. Instead of applying the indicator function of the feature map $A^{(l)}$ computed during the forward pass, they directly applied ReLU to the back-propagated values $R^{(l+1)}=\frac{\partial{o_c}}{\partial{A^{(l+1)}}}$, which corresponds to multiplying it by the indicator function $\mathbb{1}_{R^{(l+1)}>0}$. This ``backward deconvnet'' method allows back-propagating only the positive gradients, and, according to the authors, it results in a reconstructed image showing the part of the input image that is most strongly activating this neuron.

The guided back-propagation method (equation~\ref{eq: guided backprop}) combines the standard back-propagation (equation~\ref{eq: standard back-propagation}) with the backward deconvnet (equation~\ref{eq: deconvnet}): when back-propagating gradients through ReLU layers, a value is set to 0 if the corresponding top gradients or bottom data is negative. This adds an additional guidance to the standard back-propagation by preventing backward flow of negative gradients.

\begin{equation}\label{eq: standard back-propagation}
    R^{(l)} = \mathbb{1}_{A^{(l)}>0} * R^{(l+1)}
\end{equation}
\begin{equation}\label{eq: deconvnet}
    R^{(l)} = \mathbb{1}_{R^{(l+1)}>0} * R^{(l+1)}
\end{equation}
\begin{equation}\label{eq: guided backprop}
    R^{(l)} = \mathbb{1}_{A^{(l)}>0} * \mathbb{1}_{R^{(l+1)}>0} * R^{(l+1)}
\end{equation}

Any back-propagation procedure can be ``guided'', as it only concerns the way ReLU functions are managed during back-propagation (this is the case for example for guided Grad-CAM).

While it was initially adopted by the community, this method showed severe defects as discussed later in section \ref{sec:limitations}. 

\paragraph{CAM \& Grad-CAM}
 
In this setting, attribution maps are computed at the level of a feature map produced by a convolutional layer, and then upsampled to be overlaid and compared with the input. The first method, class activation maps (CAM) was proposed by Zhou et al.~\cite{zhouLearningDeepFeatures2015}, and can be only applied to CNNs with the following specific architecture:
\begin{enumerate}
    \item a series of convolutions associated with activation functions and possibly pooling layers. These convolutions output a feature map $A$ with $N$ channels,
    \item a global average pooling that extracts the mean value of each channel of the feature map produced by the convolutions,
    \item a single fully-connected layer.
\end{enumerate}
The CAM corresponding to $o_c$ will be the mean of the channels of the feature map produced by the convolutions, weighted by the weights $w_{kc}$ learned in the fully-connected layer 
\begin{equation}
    S_c = \sum_{k=1}^N w_{kc} * A_k \enspace .
\end{equation}
This map has the same size as $A_k$, which might be smaller than the input if the convolutional part performs downsampling operations (which is very often the case). Then, the map is upsampled to the size of the input to be overlaid on the input.

\begin{figure}
    \centering
    \includegraphics{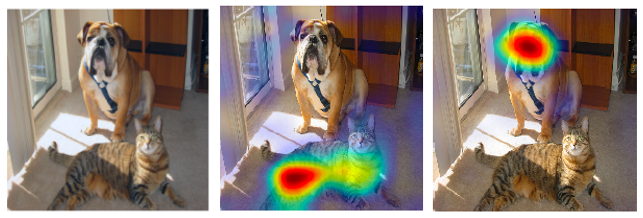}
    \caption[Grad-CAM explanations highlighting two different objects in an image.]{Grad-CAM explanations highlighting two different objects in an image. (A) the original image, (B) the explanation based on the ``dog'' node, (C) the explanation based on the ``cat'' node. \\ \textcopyright 2017 IEEE. Reprinted, with permission, from \cite{selvarajuGradCAMVisualExplanations2017}.}
    \label{fig:selvaraju_gradcam}
\end{figure}

Selvaraju et al.~\cite{selvarajuGradCAMVisualExplanations2017} proposed an extension of CAM that can be applied to any architecture: Grad-CAM (illustrated on Figure~\ref{fig:selvaraju_gradcam}). As in CAM, the attribution map is a linear combination of the channels of a feature map computed by a convolutional layer. But, in this case, the weights of each channel are computed using gradient back-propagation
\begin{equation}
    \alpha_{kc} = \frac{1}{\lvert \mathcal{U} \rvert} \sum_{u \in \mathcal{U}} \frac{\partial {o_c}}{\partial A_{k}(u)} \enspace .
\end{equation}
The final map is then the linear combination of the feature maps weighted by the coefficients. A ReLU activation is then applied to the result to only keep the features that have a positive influence on class $c$ 
\begin{equation}
    S_c = ReLU(\sum_{k=1}^N \alpha_{kc} * A_k) \enspace .
\end{equation}
Similarly to CAM, this map is then upsampled to the input size. 

Grad-CAM can be applied to any feature map produced by a convolution, but in practice the last convolutional layer is very often chosen. The authors argue that this layer is ``the best compromise between high-level semantics and detailed spatial information'' (the latter is lost in fully-connected layers, as the feature maps are flattened).

Because of the upsampling step, CAM and Grad-CAM produce maps that are more human-friendly because they contain more connected zones, contrary to other attribution maps obtained with gradient back-propagation that can look very scattered. However, the smallest the feature maps $A_k$, the blurrier they are, leading to a possible loss of interpretability.  

\subsubsection{Relevance back-propagation}
\label{subsubsec: Relevance BP}

Instead of back-propagating gradients to the level of the input or of the last convolutional layer, Bach et al.~\cite{bachPixelWiseExplanationsNonLinear2015} proposed to back-propagate the score obtained by a class $c$, which is called the relevance. This score corresponds to $o_c$ after some postprocessing (for example softmax), as its value must be positive if class $c$ was identified in the input. At the end of the back-propagation process, the goal is to find the relevance $R_u$ of each feature $u$ of the input (for example, of each pixel of an image) such that $o_c = \sum_{u \in \mathcal{U}} R_u$. 

In their paper, Bach et al.~\cite{bachPixelWiseExplanationsNonLinear2015} take the example of a fully-connected function defined by a matrix of weights $w$ and a bias $b$ at layer $l+1$. The value of a node $v$ in feature map $A^{(l+1)}$ is computed during the forward pass by the given formula:
\begin{equation}
    A^{(l+1)}(v) = b + \sum_{u \in \mathcal{U}} w_{uv} A^{(l)}(u)
\end{equation}

During the back-propagation of the relevance, $R^{(l)}(u)$, the value of the relevance at the level of the layer $l+1$, is computed according to the values of the relevance $R^{(l+1)}(v)$ which are distributed according to the weights $w$ learnt during the forward pass and the values of $A^{(l)}(v)$:
\begin{equation}
    R^{(l)}(u) = \sum_{v\in \mathcal{V}} R^{(l+1)}(v) \frac{A^{(l)}(u) w_{uv}}{\sum\limits_{u' \in \mathcal{U}} A^{(l)}(u') w_{u'v}} \enspace .
\end{equation}

The main issue of the method comes from the fact that the denominator may become (close to) zero, leading to the explosion of the relevance back-propagated. Moreover, it was shown by~ \cite{shrikumarNotJustBlack2017a} that when all activations are piece-wise linear (such as ReLU or leaky ReLU) the layer-wise relevance (LRP) method reproduces the output of gradient$\odot$input, questioning the usefulness of the method.

\begin{figure}
    \centering
    \includegraphics[width=\textwidth]{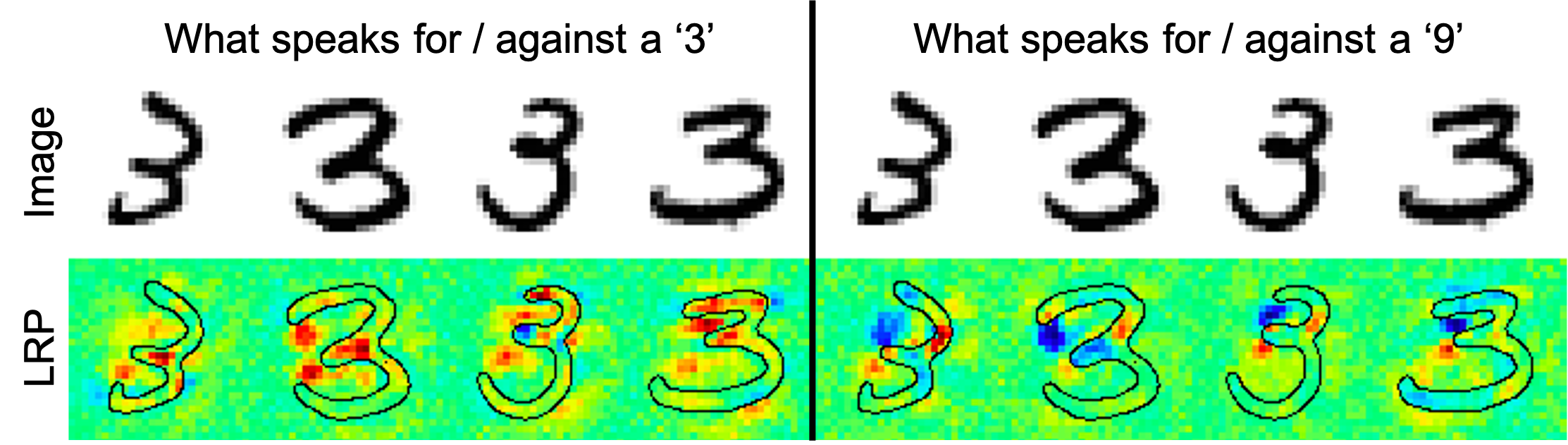}
    \caption[LRP attribution maps explaining the decision of a neural network trained on MNIST.]{LRP attribution maps explaining the decision of a neural network trained on MNIST.\\
    \textcopyright 2017 IEEE. Reprinted, with permission, from  \cite{samekEvaluatingVisualizationWhat2017}.}
    \label{fig:samek_LRP}
\end{figure}

This is why Samek et al.~\cite{samekEvaluatingVisualizationWhat2017} proposed two variants of the standard-LRP method~ \cite{bachPixelWiseExplanationsNonLinear2015}. Moreover they describe the behavior of the back-propagation in other layers than the linear ones (the convolutional one following the same formula as the linear). They illustrated their method with a neural network trained on MNIST (see Figure~\ref{fig:samek_LRP}). To simplify the equations in the following paragraphs, we now denote the weighted activations as $z_{uv} = A^{(l)}(u) w_{uv}$.

\paragraph{$\epsilon$-rule}

The $\epsilon$-rule integrates a parameter $\epsilon > 0$, used to avoid numerical instability. Though it avoids the case of a null denominator, this variant breaks the rule of relevance conservation across layers

\begin{equation}
    R^{(l)}(u) = \sum_{v \in \mathcal{V}} R^{(l+1)}(v) \frac{z_{uv}}{\sum\limits_{u' \in \mathcal{U}} z_{u'v} + \epsilon \times sign\left(\sum\limits_{u' \in \mathcal{U}} z_{u'v}\right)} \enspace .
\end{equation}

\paragraph{$\beta$-rule}

The $\beta$-rule keeps the conservation of the relevance by treating separately the positive weighted activations $z^+_{uv}$ from the negative ones $z^-_{uv}$

\begin{equation}
    R^{(l)}(u) = \sum_{v \in \mathcal{V}} R^{(l+1)}(v) \left((1 + \beta)\frac{z^+_{uv}}{\sum\limits_{u' \in \mathcal{U}} z^+_{u'v}} - \beta \frac{z^-_{uv}}{\sum\limits_{u' \in \mathcal{U}} z^-_{u'v}}\right) \enspace .
\end{equation}
Though these two LRP variants improve the numerical stability of the procedure, they imply to choose the values of parameters that may change the patterns in the obtained saliency map.

\paragraph{Deep Taylor decomposition}

Deep Taylor decomposition \citep{montavonExplainingNonlinearClassification2017} was proposed by the same team as the one that proposed the original LRP method and its variants. It is based on similar principles as LRP: the value of the score obtained by a class $c$ is back-propagated, but the back-propagation rule is based on first-order Taylor expansions.

The back-propagation from node $v$ in at the level of $R^{(l+1)}$ to $u$ at the level of $R^{(l)}$ can be written
\begin{equation}
     R^{(l)}(u) = \sum_{v \in \mathcal{V}} \frac{\partial R^{(l+1)}(v)}{\partial A^{(l)}(u)} \Bigr|_{\substack{\tilde{A}^{(l)}(u^{(v))}}} \left( A^{(l)}(u) -  \tilde{A}^{(l)}(u^{(v))} \right) \enspace .
\end{equation}
This rule implies a root point $\tilde{A}^{(l)}(u^{(v))}$ which is close to $A^{(l)}(u)$ and meets a set of constraints depending on $v$.

\subsection{Perturbation methods}
\label{subsec:perturbations}

Instead of relying on a backward pass (from the output to the input) as in the previous section, perturbation methods rely on the difference between the value of $o_c$ computed with the original inputs and a locally perturbed input. This process is less abstract for humans than back-propagation methods as we can reproduce it ourselves: if the part of the image that is needed to find the good output is hidden, we are also not able to predict correctly. Moreover, it is model-agnostic and can be applied to any algorithm or deep learning architecture.

The main drawback of these techniques is that the nature of the perturbation is crucial, leading to different attribution maps depending on the perturbation function used. Moreover, Montavon et al.~\cite{montavonMethodsInterpretingUnderstanding2018} suggest that the perturbation rule should keep the perturbed input in the training data distribution. Indeed, if it is not the case one cannot know if the network performance dropped because of the location or the nature of the perturbation.

\subsubsection{Standard perturbation}

\begin{figure}[!tbh]
    \centering
    \includegraphics[width=\textwidth]{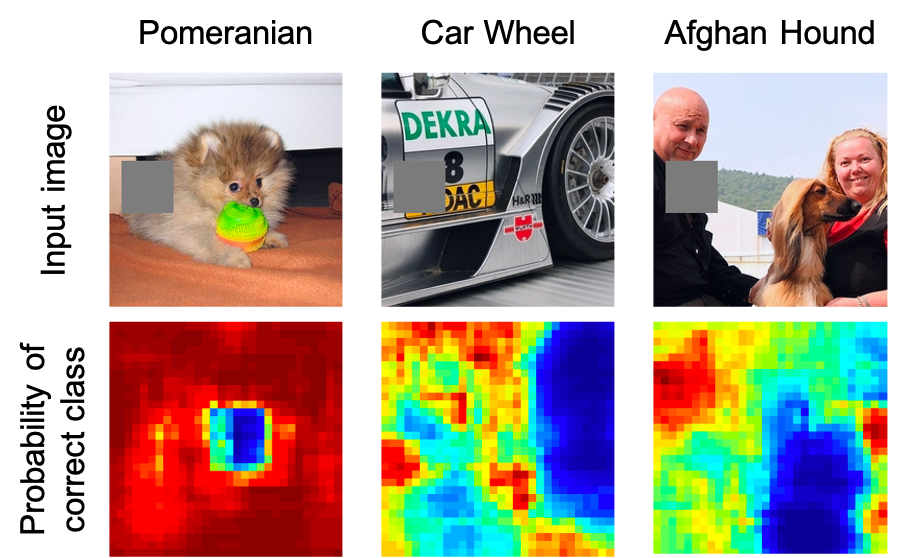}
    \caption[Attribution maps obtained with standard perturbation.]{Attribution maps obtained with standard perturbation. Here the perturbation is a gray patch covering a specific zone of the input as shown in the left column. The attribution maps (second row) display the probability of the true label: the lower the value, the most important it is for the network to correctly identify the label. This kind of perturbation takes the perturbed input out of the training distribution. \\
    Reprinted by permission from Springer Nature Customer Service Centre GmbH: Springer Nature, ECCV 2014: Visualizing and Understanding Convolutional Networks,  \citep{zeilerVisualizingUnderstandingConvolutional2014}, 2014.}
    \label{fig:standard_perturbation}
\end{figure}

Zeiler and Fergus~\cite{zeilerVisualizingUnderstandingConvolutional2014} proposed the most intuitive method relying on perturbations. This standard perturbation procedure consists in removing information locally in a specific zone of an input $X_0$ and evaluating if it modifies the output node $o_c$. The more the perturbation degrades the task performance, the more crucial this zone is for the network to correctly perform the task. To obtain the final attribution map, the input is perturbed according to all possible locations. Examples of attribution maps obtained with this method are displayed in Figure~\ref{fig:standard_perturbation}. 

As evaluating the impact of the perturbation at each pixel location is computationally expensive, one can choose not to perturb the image at each pixel location, but to skip some of them (i.e. scan the image with a stride $>$ 1). This will lead to a smaller attribution map, which needs to be upsampled to be compared to the original input (in the same way as CAM \& Grad-CAM).

However, in addition to the problem of the nature of the perturbation previously mentioned, this method presents two drawbacks:
\begin{itemize}
    \item the attribution maps depend on the size of the perturbation: if the perturbation becomes too large, the perturbation is not local anymore, if it too small it is not meaningful anymore (a pixel perturbation cannot cover a pattern),
    \item input pixels are considered independently from each other: if the result of a network relies on a combination of pixels that cannot all be covered at the same time by the perturbation, their influence may not be detected.
\end{itemize}

\subsubsection{Optimized perturbation}

To deal with these two issues, 
Fong and Vedaldi~\cite{fongInterpretableExplanationsBlack2017} proposed to optimize a perturbation mask covering the whole input. This perturbation mask $m$ has the same size as the input $X_0$. Its application is associated with a perturbation function $\Phi$ and leads to the computation of the perturbed input $X_0^m$. Its value at a coordinate $u$ reflects the quantity of information remaining in the perturbed image:
\begin{itemize}
    \item if $m(u) = 1$, the pixel at location $u$ is not perturbed and has the same value in the perturbed input as in the original input ($X_0^m(u)=X_0(u)$).
    \item if $m(u) = 0$ the pixel at location $u$ is fully perturbed and the value in the perturbed image is the one given by the perturbation function only ($X_0^m(u)=\Phi(X_0)(u)$).
\end{itemize}
This principle can be extended to any value between 0 and 1 with the a linear interpolation
\begin{equation}
    X_0^m(u)= m(u)X_0(u) + (1-m(u))\Phi(X_0)(u) \enspace .
\end{equation}
Then, the goal is to optimize this mask $m$ according to three criteria:
\begin{enumerate}
    \item the perturbed input $X_0^m$ should lead to the lowest performance possible,
    \item the mask $m$ should perturb the minimum number of pixels possible, and
    \item the mask $m$ should produce connected zones (i.e. avoid the scattered aspect of gradient maps).
\end{enumerate}
These three criteria are optimized using the following loss:
\begin{equation}
    f(X_0^m) + \lambda_1 \lVert 1 - m \rVert^{\beta_1}_{\beta_1} + \lambda_2 \lVert \nabla m \rVert^{\beta_2}_{\beta_2}
\end{equation}
with $f$ a function that decreases as the performance of the network decreases.

\begin{figure}[!tbh]
    \centering
    \includegraphics[width=\textwidth]{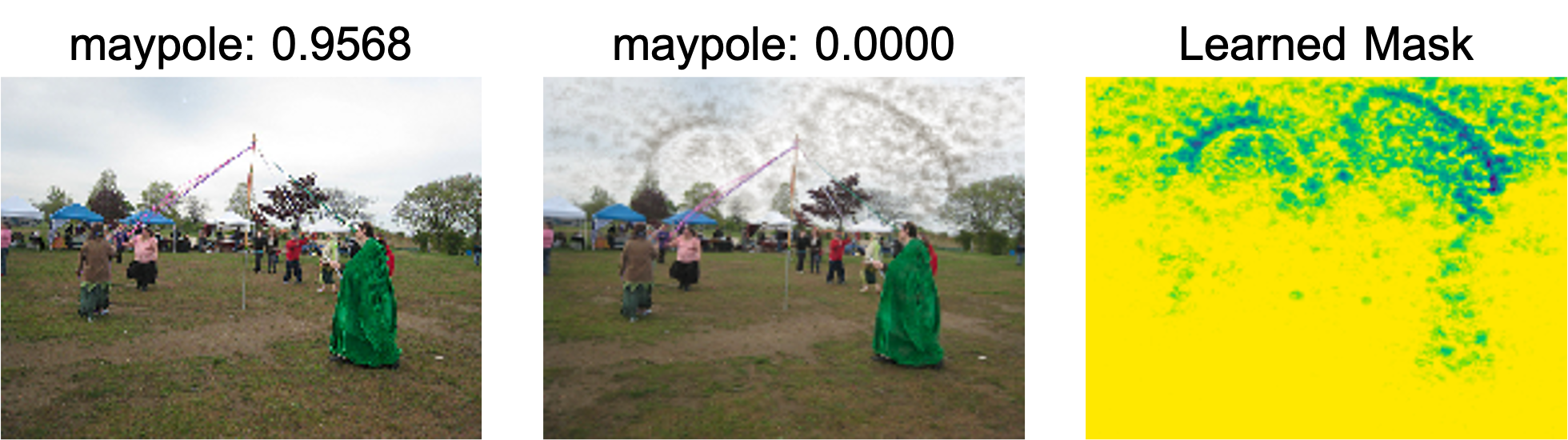}
    \caption[Example of artifacts created by optimized perturbation method.]{In this example, the network learned to classify objects in natural images. Instead of masking the maypole at the center of the image, it creates artifacts in the sky to degrade the performance of the network. \\
    \textcopyright 2017 IEEE. Reprinted, with permission, from \citep{fongInterpretableExplanationsBlack2017}.}
    \label{fig:optimized_artifacts}
\end{figure}

However, the method also presents two drawbacks:
\begin{itemize}
    \item The values of hyperparameters must be chosen ($\lambda_1$, $\lambda_2$, $\beta_1$, $\beta_2$) to find a balance between the three optimization criteria of the mask,
    \item The mask may not highlight the most important pixels of the input but instead create artifacts in the perturbed image to artificially degrade the performance of the network (see Figure~\ref{fig:optimized_artifacts}).
\end{itemize}

\subsection{Distillation}
\label{subsec:distillation}

Approaches described in this section aim at developing a transparent method to reproduce the behavior of a black-box one. Then it is possible to consider simple interpretability methods (such as weight visualization) on the transparent method instead of considering the black box.

\subsubsection{Local approximation}

\paragraph{LIME}

Ribeiro et al.~\cite{ribeiroWhyShouldTrust2016} proposed Local Interpretable Model-agnostic Explanations (LIME). This approach is:
\begin{itemize}
    \item \textbf{local}, as the explanation is valid in the vicinity of a specific input $X_0$,
    \item \textbf{interpretable}, as an interpretable model $g$ (linear model, decision tree...) is computed to reproduce the behavior of $f$ on $X_0$, and
    \item \textbf{model-agnostic}, as it does not depend on the algorithm trained.
\end{itemize}
This last property comes from the fact that the vicinity of $X_0$ is explored by sampling variations of $X_0$ that are perturbed versions of $X_0$. Then LIME shares the advantage (model agnostic) and drawback (perturbation function dependent) of perturbations methods presented in section~\ref{subsec:perturbations}. Moreover, the authors specify that, in the case of images, they group pixels of the input in $d$ super-pixels (contiguous patches of similar pixels). 

The loss to be minimized to find $g$ specific to the input $X_0$  is the following:
\begin{equation}
    \mathcal{L}(f, g, \pi_{X_0}) + \Omega(g) \enspace ,
\end{equation}
where $\pi_{X_0}$ is a function that defines the locality of $X_0$ (i.e. $\pi_{X_0}(X)$ decreases as $X$ becomes closer to $X_0$),
$\mathcal{L}$ measures how unfaithful $g$ is in approximating $f$ according $\pi_{X_0}$, and
$\Omega$ is a measure of the complexity of $g$.

Ribeiro et al.~\cite{ribeiroWhyShouldTrust2016} limited their search to sparse linear models, however other assumptions could be made on $g$.

$g$ is not applied to the input directly but to a binary mask $m \in \{0, 1\}^d$ that transforms the input $X$ in $X^m$ and is applied according to a set of $d$ super-pixels. For each super-pixel $u$:
\begin{enumerate}
    \item if $m(u) = 1$ the super-pixel $u$ is not perturbed,
    \item if $m(u) = 0$ the super-pixel $u$ is perturbed (i.e. it is grayed).
\end{enumerate}
They used $\pi_{X_0}(X) = \exp{\frac{(X - X_0)^2}{\sigma^2}}$ and $\mathcal{L}(f, g, \pi_{X_0}) = \sum_{m} \pi_{X_0}(X_0^m) * (f(X_0^m) - g(m))^2$. Finally $\Omega(g)$ is the number of non-zero weights of $g$, and its value is limited to $K$. This way they select the $K$ super-pixels in $X_0$ that best explain the algorithm result $f(X_0)$.


\paragraph{SHAP}

Lundberg and Lee~\cite{lundbergUnifiedApproachInterpreting2017a} proposed SHAP (SHapley Additive exPlanations), a theoretical framework that encompasses several existing interpretability methods, including LIME. In this framework each of the $N$ features (again, super-pixels for images) is associated with a coefficient $\phi$ that denotes its contribution to the result. The contribution of each feature is evaluated by perturbing the input $X_0$ with a binary mask $m$ (see paragraph on LIME). Then the goal is to find an interpretable model $g$ specific to $X_0$, such that
\begin{equation}
    g(m) = \phi_0 + \sum_1^N{\phi_i m_i}
\end{equation}
with $\phi_0$ being the output when the input is fully perturbed.

The authors look for an expression of $\phi$ that respects three properties:
\begin{itemize}
    \item \textbf{Local accuracy}\quad $g$ and $f$ should match in the vincinity of $X_0$: $g(m) = f(X_0^m)$.
    \item \textbf{Missingness}\quad Perturbed features should not contribute to the result: $m_i = 0 \rightarrow \phi_i = 0$.
    \item \textbf{Consistency}\quad Let's denote as $m \setminus i$ the mask $m$ in which $m_i = 0$. For any two models $f^1$ and $f^2$, if
    $f^1(X_0^{m}) - f^1(X_0^{m \setminus i}) \ge f^2(X_0^{m}) - f^2(X_0^{m \setminus i})$,
    then for all $m \in \{0, 1\}^N$
    $\phi^1_i \ge \phi^2_i$ ($\phi^k$ are the coefficients associated with model $f^k$).
\end{itemize}

Lundberg and Lee~\cite{lundbergUnifiedApproachInterpreting2017a} show that only one expression is possible for the coefficients $\phi$, which can be approximated with different algorithms:

\begin{equation}
    \phi_i = \sum_{m \in \{0, 1\}^N} \frac{|m|! (N - |m| - 1)!}{N!} \left[ f(X_0^{m}) - f(X_0^{m \setminus i}) \right] \enspace .
\end{equation}

\subsubsection{Model translation}

Contrary to local approximation, which provides an explanation according to a specific input $X_0$, model translation consists in finding a transparent model that reproduces the behavior of the black-box model on the whole data set.

As it was rarely employed in neuroimaging frameworks, this section only discusses the distillation to decision trees proposed in \cite{frosstDistillingNeuralNetwork2017} (preprint). For a more extensive review of model translation methods, we refer the reader to  \cite{xieExplainableDeepLearning2020}.

\begin{figure}
    \centering
    \includegraphics[width=\textwidth]{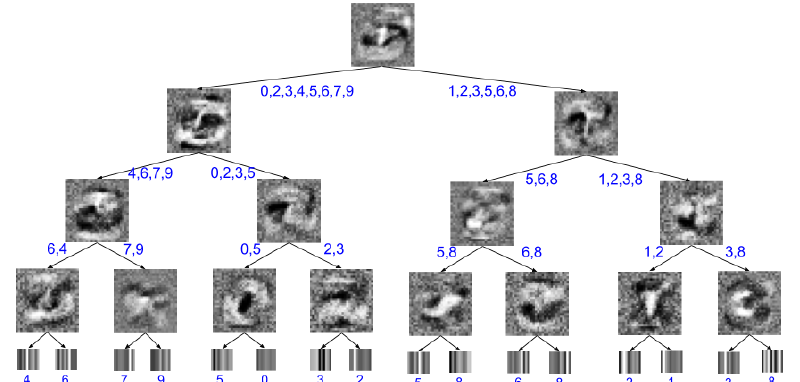}
    \caption[Visualization of a soft decision tree trained on MNIST.]{Visualization of a soft decision tree trained on MNIST. \\
    Adapted from \cite{frosstDistillingNeuralNetwork2017}. Permission to reuse was kindly granted by the authors.}
    \label{fig: frosst_tree}
\end{figure}

After training a machine learning system $f$, a binary decision tree $g$ is trained to reproduce its behavior. This tree is trained on a set of inputs $X$, and each inner node $i$ learns a matrix of weights $w_i$ and biases $b_i$. The forward pass of $X$ in the node $i$ of the tree is as follows: if $sigmoid(w_iX + b_i) > 0.5$, then the right leaf node is chosen, else the left leaf node is chosen. After the end of the decision tree's training, it is possible to visualize at which level which classes were separated to better understand which classes are similar for the network. It is also possible to visualize the matrices of weights learned by each inner node to identify patterns learned at each class separation. An illustration of this distillation process, on the MNIST data set (hand-written digits), can be found in Figure~\ref{fig: frosst_tree}.

\subsection{Intrinsic}
\label{subsec:intrisinc}

Contrary to the previous sections in which interpretability methods could be applied to (almost) any network after the end of the training procedure, the following methods require to design the framework before the training phase, as the interpretability components and the network are trained simultaneously. In the papers presented in this section~\citep{xuShowAttendTell2016, wangResidualAttentionNetwork2017a, baMultipleObjectRecognition2015}, the advantages of these methods are dual: they improve both the interpretability and performance of the network. However, the drawback is that they have to be implemented before training the network, then they cannot be applied in all cases.

\subsubsection{Attention modules}

Attention is a concept in machine learning that consists in producing an attribution map from a feature map and using it to improve learning of another task (such as classification, regression, reconstruction...) by making the algorithm focus on the part of the feature map highlighted by the attribution map.

\begin{figure}
    \centering
    \includegraphics[width=\textwidth]{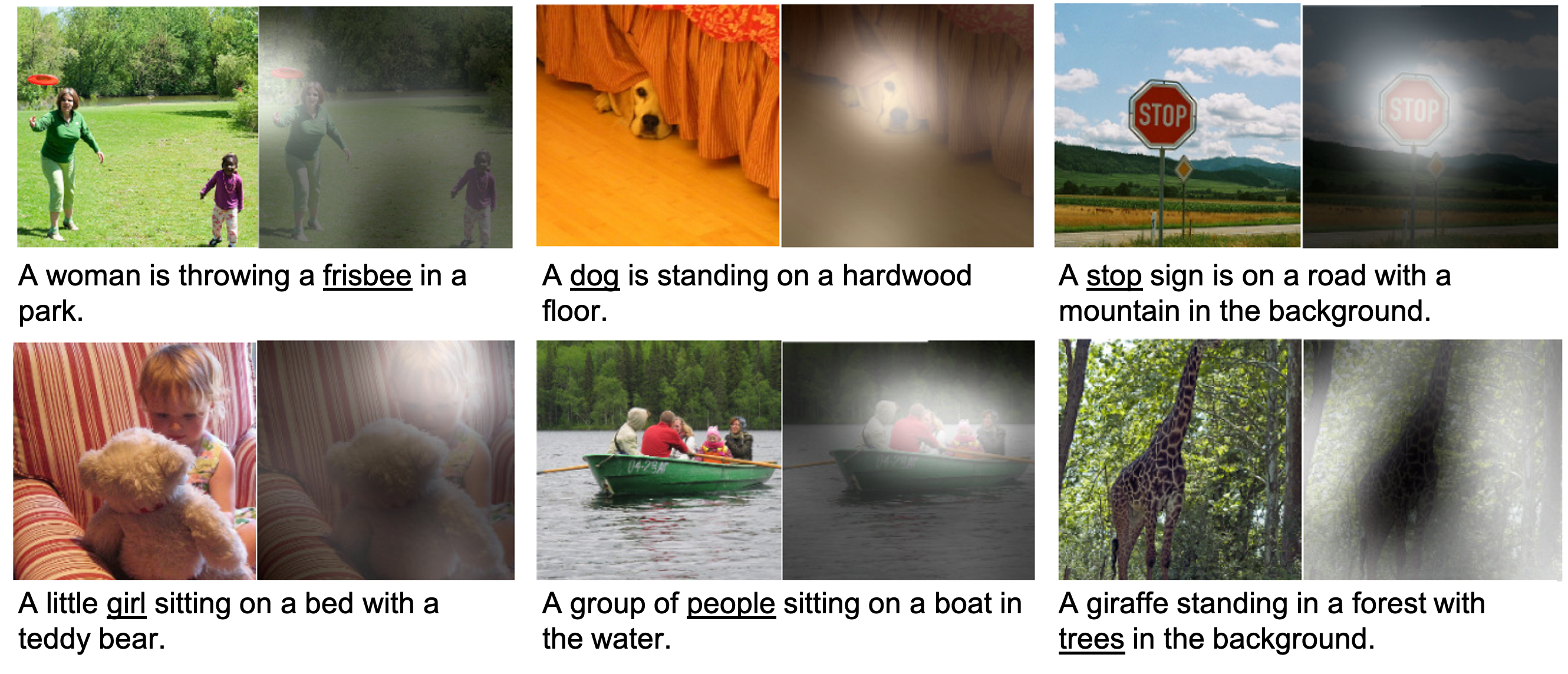}
    \caption[Attribution maps obtained with attention modules.]{Examples of images correctly captioned by the network. The focus of the attribution map is highlighted in white and the associated word in the caption is underlined. \\
    Adapted from \citep{xuShowAttendTell2016}. Permission to reuse was kindly granted by the authors.}
    \label{fig:attention_lstm}
\end{figure}

In the deep learning domain, we take as reference \cite{xuShowAttendTell2016}, in which a network is trained to produce a descriptive caption of natural images. This network is composed of three parts: 
\begin{enumerate}
    \item a convolutional encoder that reduces the dimension of the input image to the size of the feature maps $A$,
    \item an attention module that generates an attribution map $S_t$ from $A$ and the previous hidden state of the long short-term memory (LSTM) network,
    \item an LSTM decoder that computes the caption from its previous hidden state, the previous word generated, $A$ and $S_t$.
\end{enumerate}
As $S_t$ is of the same size as $A$ (smaller than the input), the result is then upsampled to be overlaid on the input image. As one attribution map is generated per word generated by the LSTM, it is possible to know where the network focused when generating each word of the caption (see Figure~\ref{fig:attention_lstm}). 
In this example, the attribution map is given to a LSTM, which uses it to generate a context vector $z_t$ by applying a function $\phi$ to $A$ and $S_t$. 

More generally in CNNs, the point-wise product of the attribution map $S$ and the feature map $A$ is used to generate the refined feature map $A'$ which is given to the next layers of the network. Adding an attention module implies to make new choices for the architecture of the model: its location (on lower or higher feature maps) may impact the performance of the network. Moreover, it is possible to stack several attention modules along the network, as it was done in \cite{wangResidualAttentionNetwork2017a}.

\subsubsection{Modular Transparency}

Contrary to the studies of the previous sections, the frameworks of these categories are composed of several networks (modules) that interact with each other. Each module is a black box, but the transparency of the function, or the nature of the interaction between them, allows understanding how the system works globally and extracting interpretability metrics from it.


A large variety of setups can be designed following this principle, and it is not possible to draw a more detailed general rule for this section. We will take the example described in~\cite{baMultipleObjectRecognition2015}, which was adapted to neuroimaging data (see Section~\ref{sec:application_intrinsic}), to illustrate this section, though it may not be representative of all the aspects of modular transparency.

\begin{figure}
    \centering
    \includegraphics[width=\textwidth]{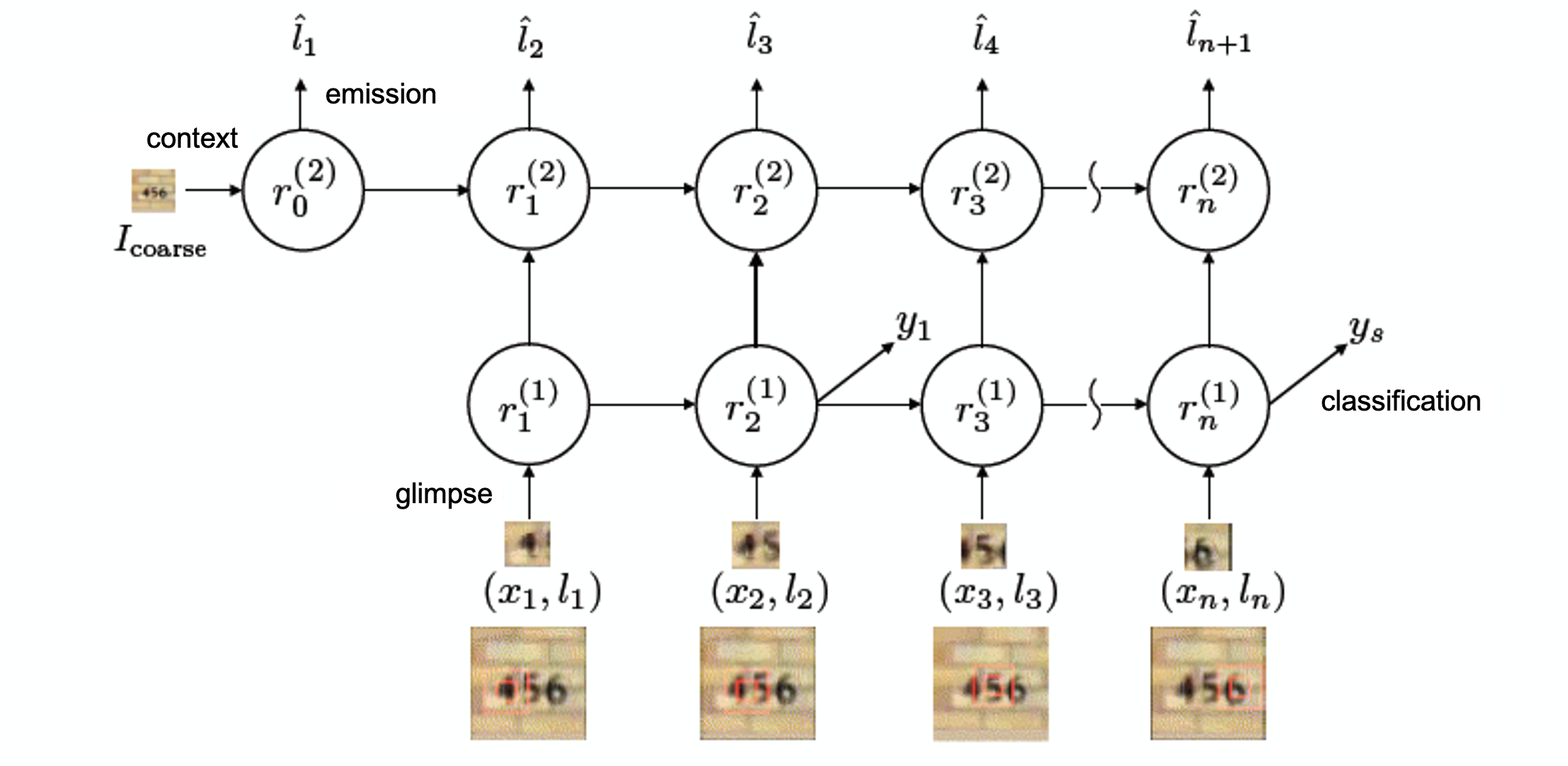}
    \caption[Framework with modular transparency browsing an image to compute the output at the global scale.]{Framework with modular transparency browsing an image to compute the output at the global scale. \\
    Adapted from \citep{baMultipleObjectRecognition2015}. Permission to reuse was kindly granted by the authors.}
    \label{fig: ba_modular}
\end{figure}

Ba et al.~\cite{baMultipleObjectRecognition2015} proposed a framework (illustrated in Figure~\ref{fig: ba_modular}) to perform the analysis of an image in the same way as a human, by looking at successive relevant locations in the image. To perform this task, they assemble a set of networks that interact together:
\begin{itemize}
    \item \textbf{Glimpse network}\quad This network takes as input a patch of the input image and the location of its center to output a context vector that will be processed by the recurrent network. Then this vector conveys information on the main features in a patch and its location.
    \item \textbf{Recurrent network}\quad This network takes as input the successive context vectors and update its hidden state that will be used to find the next location to look at and to perform the learned task at the global scale (in the original paper a classification of the whole input image).
    \item \textbf{Emission network}\quad This network takes as input the current state of the recurrent network and outputs the next location to look at. This will allow computing the patch that will feed the glimpse network.
    \item \textbf{Context network}\quad This network takes as input the whole input at the beginning of the task and outputs the first context vector to initialize the recurrent network.
    \item \textbf{Classification network}\quad This network takes as input the current state of the recurrent network and outputs a prediction for the class label.
\end{itemize}
The global framework can be seen as interpretable as it is possible to review the successive processed locations.

\subsection{Interpretability metrics}
\label{sec:evaluation_metrics}

To evaluate the reliability of the methods presented in the previous sections, one cannot only rely on qualitative evaluation. This is why interpretability metrics that evaluate attribution maps were proposed.
These metrics may evaluate different properties of attribution maps.
\begin{itemize}
    \item \textbf{Fidelity} evaluates if the zones highlighted by the map influence the decision of the network.
    \item \textbf{Sensitivity} evaluates how the attribution map changes according to small changes in the input $X_0$.
    \item \textbf{Continuity} evaluates if two close data points lead to similar attribution maps.
\end{itemize}
In the following, $\Gamma$ is an interpretability method computing an attribution map $S$ of the black-box network $f$ and an input $X_0$.

\subsubsection{(In)fidelity}

Yeh et al.~\cite{yehFidelitySensitivityExplanations2019} proposed a measure of infidelity of $\Gamma$ based on perturbations applied according to a vector $m$ of the same shape as the attribution map $S$. The explanation is infidel if perturbations applied in zones highlighted by $S$ on $X_0$ leads to negligible changes in $f(X_0^m)$ or, on the contrary, if perturbations applied in zones not highlighted by $S$ on $X_0$ lead to significant changes in $f(X_0^m)$. The associated formula is

\begin{equation}
    \text{INFD}(\Gamma, f, X_0) = \mathbb{E}_{m} \left[ \sum_{i}\sum_{j}m_{ij} \Gamma(f, X_0)_{ij} - (f(X_0) - f(X_0^m))^2 \right] \enspace .
\end{equation}

\subsubsection{Sensitivity}

Yeh et al.~\cite{yehFidelitySensitivityExplanations2019} also gave a measure of sensitivity. As suggested by the definition, it relies on the construction of attribution maps according to inputs similar to $X_0$: $\tilde{X_0}$. As changes are small, sensitivity depends on a scalar $\epsilon$ set by the user, which corresponds to the maximum difference allowed between $X_0$ and $\tilde{X_0}$. Then sensitivity corresponds to the following formula:

\begin{equation}
    \text{SENS}_{\text{max}}(\Gamma, f, X_0, \epsilon) = \max_{\lVert \tilde{X_0} - X_0 \rVert \le \epsilon} \lVert \Gamma(f, \tilde{X_0}) - \Gamma(f, X_0) \rVert \enspace .
\end{equation}

\subsubsection{Continuity}

Continuity is very similar to sensitivity, except that it compares different data points belonging to the input domain $\mathcal{X}$, whereas sensitivity may generate similar inputs with a perturbation method.
This measure was introduced in \cite{montavonMethodsInterpretingUnderstanding2018} and can be computed using the following formula:

\begin{equation}
    \text{CONT}(\Gamma, f, \mathcal{X}) = \max_{X_1, X_2 \in \mathcal{X}~\& ~X_1 \neq  X_2} \frac{\lVert \Gamma(f, X_1) - \Gamma(f, X_2) \rVert_1}{\lVert X_1 - X_2 \rVert_2} \enspace .
\end{equation}

\vspace{1cm}

As these metrics rely on perturbation, they are also influenced by the nature of the perturbation and may lead to different results, which is a major issue (see Section~\ref{sec:limitations}).
Other metrics were also proposed and depend on the task learned by the network: for example in the case of a classification, statistical tests can be conducted between saliency maps of different classes to assess whether they differ according to the class they explain.

\section{Application of interpretability methods to neuroimaging data} 
\label{sec:section3}


In this section, we provide a non-exhaustive review of applications of interpretability methods to neuroimaging data. In most cases, the focus of articles is prediction/classification rather than the interpretability method, which is just seen as a tool to analyze the results. Thus, authors do not usually motivate their choice of an interpretability method. Another key consideration here is the spatial registration of brain images, which enables having brain regions roughly at the same position between subjects. This technique is of paramount importance as attribution maps computed for registered images can then be averaged or used to automatically determine the most important brain areas, which would not be possible with unaligned images. 
All the studies presented in this section are summarized in Table~\ref{tab:section3}.

This section ends with the presentation of benchmarks conducted in the literature to compare different interpretability methods in the context of brain disorders.

\input{studies_table_wo_preprocessing}

\subsection{Weight visualization applied to neuroimaging}
\label{sec:application_weights}

\begin{figure}[!tbh]
    \centering
    \includegraphics[width=0.5\textwidth]{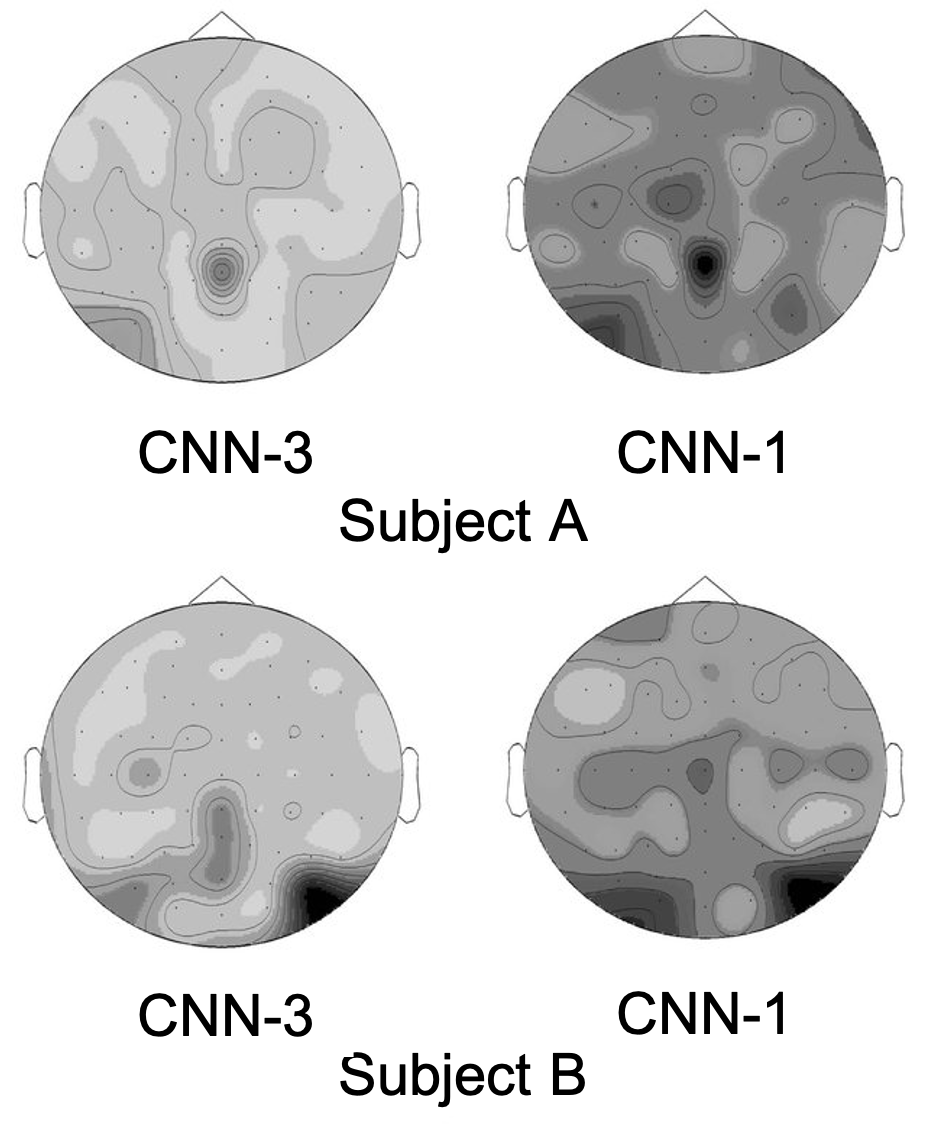}
    \caption[Relative importance of the electrodes for signal detection in EEG using CNN weight visualization]{Relative importance of the electrodes for signal detection in EEG using two different architectures (CNN-1 and CNN-3) and two subjects (A and B) using CNN weight visualization. Dark values correspond to weights with a high absolute value while white values correspond to weights close to 0.\\
    \textcopyright 2011 IEEE. Reprinted, with permission, from \citep{cecottiConvolutionalNeuralNetworks2011}.}
    \label{fig:cecotti_weights}
\end{figure}

As the focus of this chapter is on non-transparent models, such as deep learning ones, weight visualization was only rarely found. However, this was the method chosen by 
Cecotti and Gr\"{a}ser~\cite{cecottiConvolutionalNeuralNetworks2011}, who developed a CNN architecture adapted to weight visualization to detect P300 signals in electroencephalograms (EEG). The input of this network is a matrix with rows corresponding to the 64 electrodes and columns to 78 time points. The two first layers of the networks are convolutions with rectangular filters: the first filters (size 1$\times$64) combines the electrodes, whereas the second ones (13$\times$1) find time patterns. Then, it is possible to retrieve a coefficient per electrode by summing the weights associated with this electrode across the different filters, and to visualize the results in the  electroencephalogram space as show in Figure~\ref{fig:cecotti_weights}.

\subsection{Feature map visualization applied to neuroimaging}
\label{sec:application_FM}

Contrary to the limited application of weight visualization, there is an extensive literature about leveraging individual feature maps and latent spaces to better understand how models work. This goes from the visualization of these maps or their projections \citep{ohClassificationVisualizationAlzheimer2019, abrolDeepResidualLearning2020, biffiExplainableAnatomicalShape2020}, to the analysis of neuron behavior \citep{martinez-murciaStudyingManifoldStructure2020, lemingEnsembleDeepLearning2020}, through sampling in latent spaces \citep{biffiExplainableAnatomicalShape2020}.

\begin{figure}[!tbh]
    \centering
    \includegraphics[width=0.8\textwidth]{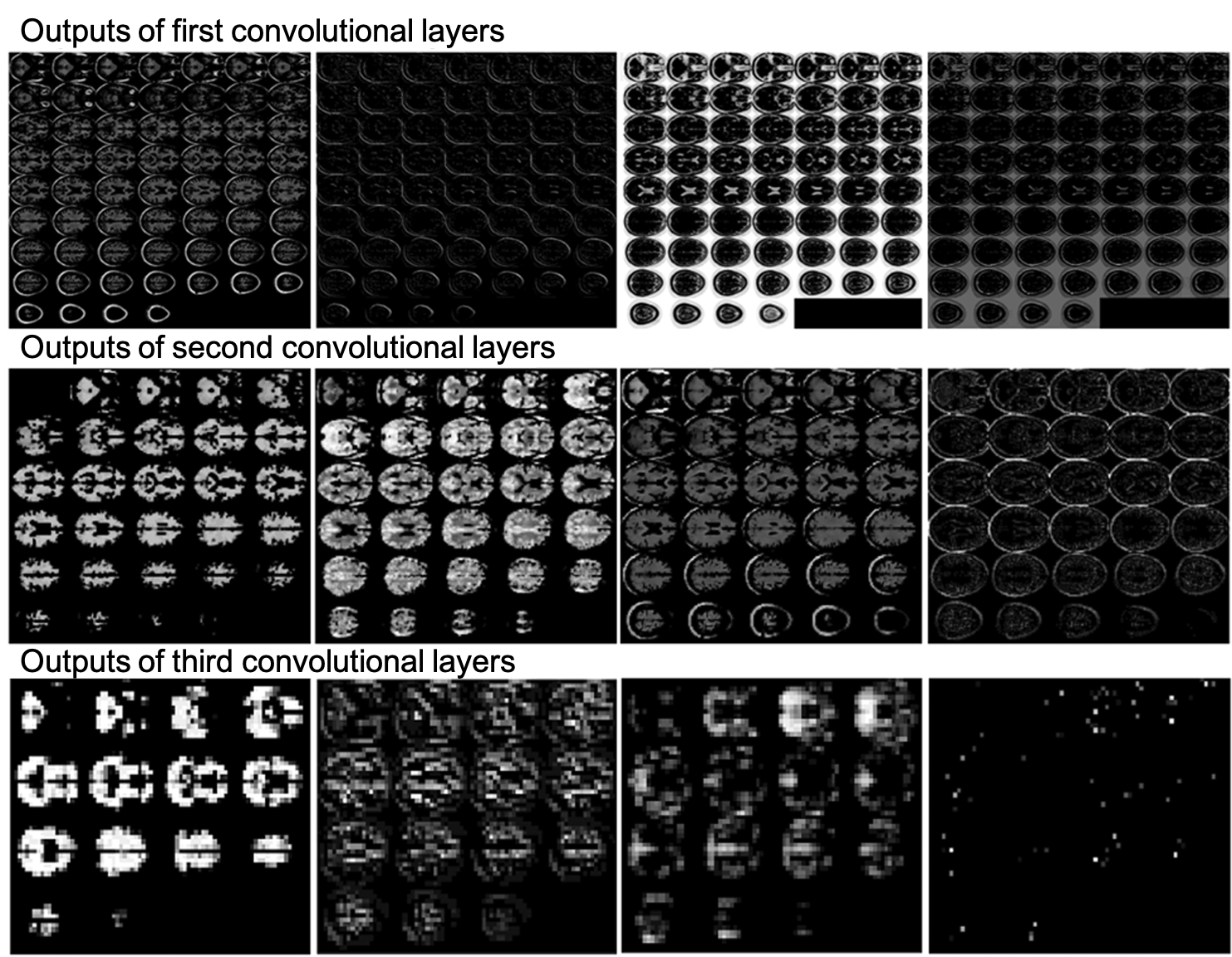}
    \caption[Representation of a selection of feature maps.]{Representation of a selection of feature maps (outputs of 4 filters on 10 for each layer) obtained for a single individual. \\
    Adapted from \citep{ohClassificationVisualizationAlzheimer2019} (CC BY 4.0).}
    \label{fig: oh_FM}
\end{figure}

Oh et al.~\cite{ohClassificationVisualizationAlzheimer2019} displayed the feature maps associated with the convolutional layers of CNNs trained for various Alzheimer's disease status classification tasks (Figure~\ref{fig: oh_FM}). In the first two layers, the extracted features were similar to white matter, cerebrospinal fluid and skull segmentations, while the last layer showcased sparse, global and nearly binary patterns. They used this example to emphasize the advantage of using CNNs to extract very abstract and complex features rather than using custom algorithms for features extraction~ \cite{ohClassificationVisualizationAlzheimer2019}.

Another way to visualize a feature map is to project it in a two or three-dimensional space to understand how it is positioned with respect to other feature maps. Abrol et al.~\cite{abrolDeepResidualLearning2020} projected the features obtained after the first dense layer of a ResNet architecture onto a two-dimensional space using the classical t-distributed stochastic neighbor embedding (t-SNE) dimensionality reduction technique. For the classification task of Alzheimer's disease statuses, they observed that the projections were correctly ordered according to the disease severity, supporting the correctness of the model~\cite{abrolDeepResidualLearning2020}. They partitioned these projections into three groups: Far-AD (more extreme Alzheimer's Disease patients), Far-CN (more extreme Cognitively Normal participants) and Fused (a set of images at the intersection of AD and CN groups). Using a t-test, they were able to detect and highlight voxels presenting significant differences between groups.

\begin{figure}[!tbh]
    \centering
    \includegraphics[width=0.95\textwidth]{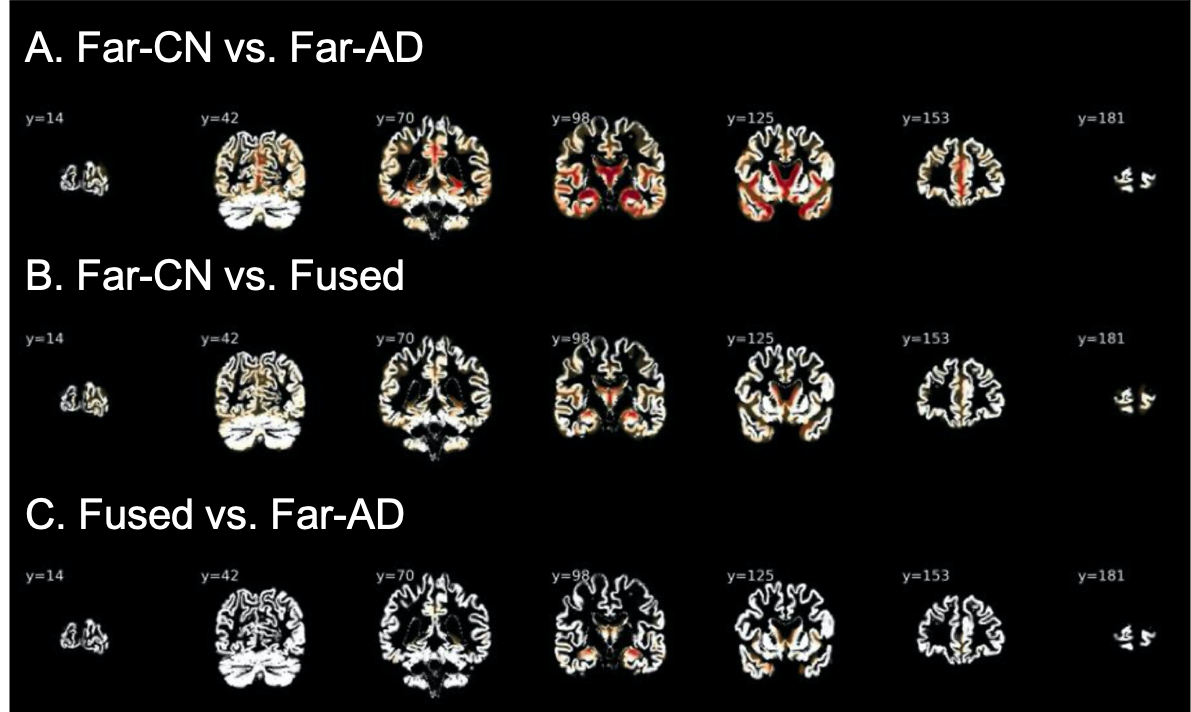}
    \caption[Difference in neuroimaging space between groups defined thanks to t-SNE projection.]{Difference in neuroimaging space between groups defined thanks to t-SNE projection. Voxels showing significant differences post false discovery rate (FDR) correction (p \textless 0.05) are highlighted. \\
    Reprinted from Journal of Neuroscience Methods, 339, \citep{abrolDeepResidualLearning2020}, 2020, with permission from Elsevier.}
    \label{fig: abrol_FM}
\end{figure}

Biffi et al.~\cite{biffiExplainableAnatomicalShape2020} not only used feature map visualization, but also sampled the feature space. Indeed, they trained a ladder variational autoencoder framework to learn hierarchical latent representations of 3D hippocampal segmentations of control subjects and Alzheimer’s disease patients. A multi-layer perceptron was jointly trained on top of the highest two-dimensional latent space to classify anatomical shapes. While lower spaces needed a dimensionality reduction technique (i.e. t-SNE), the highest latent space could directly be visualized, as well as the anatomical variability it captured in the initial input space, by leveraging the generative process of the model. This sampling enabled an easy visualization and quantification of the anatomical differences between each class.

Finally, it may be very informative to better understand the behavior of neurons and what they are encoding. After training deep convolutional autoencoders to reconstruct MR images, segmented gray matter maps and white matter maps, Martinez-Murcia et al.~\cite{martinez-murciaStudyingManifoldStructure2020} computed correlations between each individual hidden neuron value and clinical information (e.g. age, mini-mental state examination) which allowed them to determine to which extent this information was encoded in the latent space. This way they determined which clinical data was the most strongly associated.
Using a collection of nine different MRI data sets, Leming et al.~\cite{lemingEnsembleDeepLearning2020} trained CNNs for various classification tasks (autism vs typically developing, male vs female and task vs rest). They computed a diversity coefficient for each filter of the second layer based on its output feature map. They counted how many different data sets maximally activated each value of this feature map: if they were mainly activated by one source of data the coefficient would be close to 0, whereas if they were activated by all data sets it would be close to 1. This allows assessing the layer stratification, i.e. to understand if a given filter was mostly maximally activated by one phenotype or by a diverse population. They found out that a few filters were only maximally activated by images from a single MRI data set, and that the diversity coefficient was not normally distributed across filters, having generally two peaks at the beginning and at the end of the spectrum, respectively exhibiting the stratification and strongly diverse distribution of the filters.

\subsection{Back-propagation methods applied to neuroimaging}
\label{sec:application_BP}

Back-propagation methods are the most popular methods to interpret models, and a wide range of these algorithms have been used to study brain disorders: standard and guided back-propagation \citep{huDeepLearningBasedClassification2021, ohClassificationVisualizationAlzheimer2019, riekeVisualizingConvolutionalNetworks2018, eitelTestingRobustnessAttribution2019, bohleLayerwiseRelevancePropagation2019}, gradient$\odot$input \citep{eitelUncoveringConvolutionalNeural2019, eitelTestingRobustnessAttribution2019, dyrbaComparisonCNNVisualization2020}, Grad-CAM \citep{burdujaAccurateEfficientIntracranial2020, dyrbaComparisonCNNVisualization2020}, guided Grad-CAM \citep{tangInterpretableClassificationAlzheimer2019}, LRP \citep{eitelUncoveringConvolutionalNeural2019, eitelTestingRobustnessAttribution2019, dyrbaComparisonCNNVisualization2020, bohleLayerwiseRelevancePropagation2019}, DeconvNet \citep{dyrbaComparisonCNNVisualization2020} and deep Taylor Decomposition \citep{dyrbaComparisonCNNVisualization2020}. 

\subsubsection{Single interpretation}

Some studies implemented a single back-propagation method, and exploited it to find which brain regions are exploited by their algorithm \citep{ohClassificationVisualizationAlzheimer2019, lemingEnsembleDeepLearning2020, huDeepLearningBasedClassification2021}, to validate interpretability methods \citep{eitelUncoveringConvolutionalNeural2019} or to provide attribution maps to physicians to improve clinical guidance \citep{burdujaAccurateEfficientIntracranial2020}.

\begin{figure}[!tbh]
    \centering
    \includegraphics[width=\textwidth]{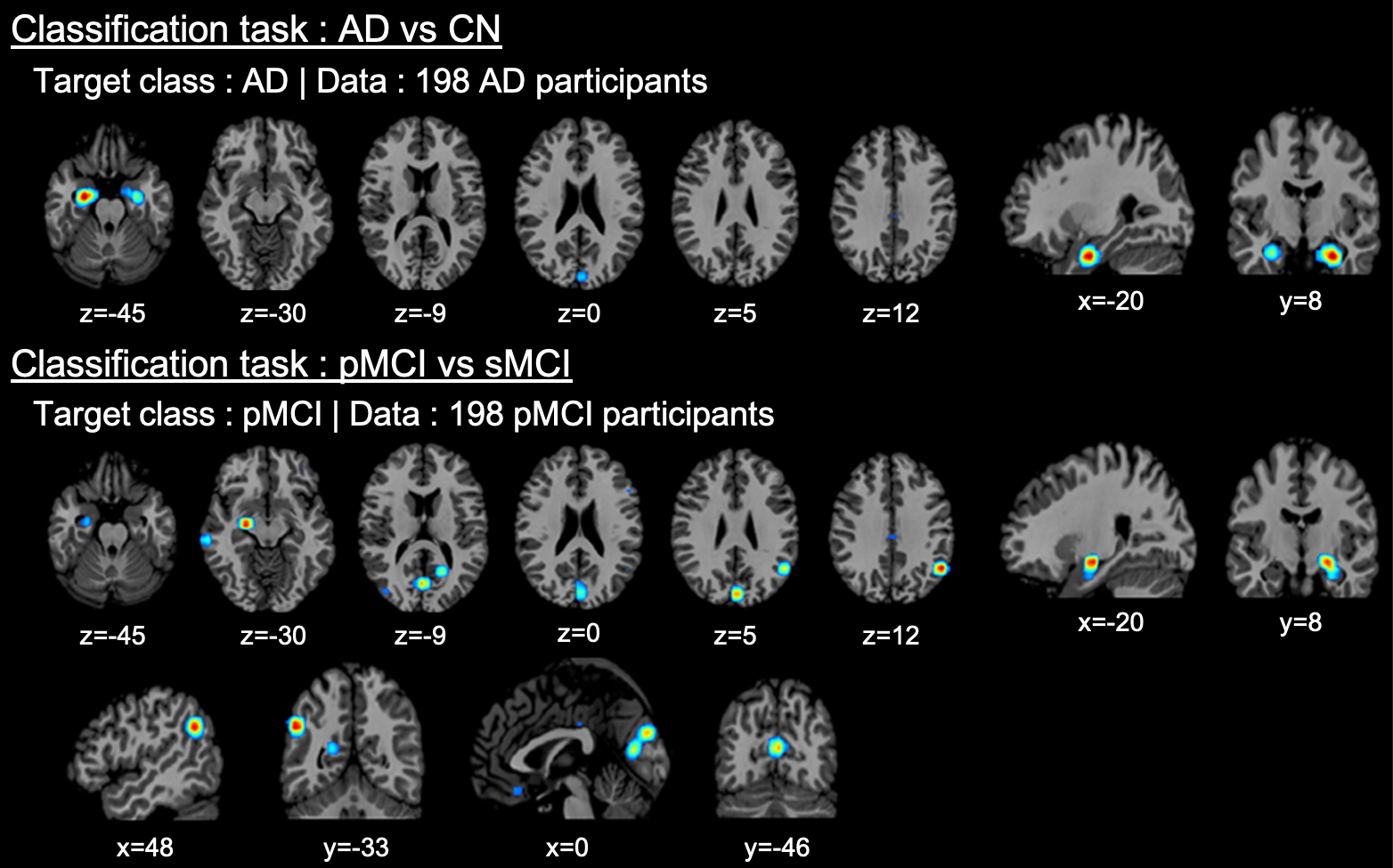}
    \caption[Distribution of discriminant regions obtained with gradient back-propagation.]{Distribution of discriminant regions obtained with gradient back-propagation in the classification of demented patients and cognitively normal participants (top part, AD vs CN) and the classification of stable and progressive mild cognitive impairment (bottom part, sMCI vs pMCI). \\
    Adapted from \citep{ohClassificationVisualizationAlzheimer2019} (CC BY 4.0).}
    \label{fig: oh_BP}
\end{figure}

Oh et al.~\cite{ohClassificationVisualizationAlzheimer2019} used the standard back-propagation method to interpret CNNs for classification of Alzheimer's disease statuses. They showed that the attribution maps associated with the prediction of the conversion of prodromal patients to dementia included more complex representations, less focused on the hippocampi, than the ones associated with classification between demented patients from cognitively normal participants (see Figure~\ref{fig: oh_BP}).
In the context of autism, Leming et al.~\cite{lemingEnsembleDeepLearning2020} used the Grad-CAM algorithm to determine the most important brain connections from functional connectivity matrices . However, the authors pointed out that without further work, this visualization method did not allow understanding the underlying reason of the attribution of a given feature: for instance, one cannot know if a set of edges is important because it is under-connected or over-connected. Finally, Hu et al.~\cite{huDeepLearningBasedClassification2021} used attribution maps produced by guided back-propagation to quantify the difference in the regions used by their network to characterize Alzheimer's disease or fronto-temporal dementia.

The goal of Eitel et al.~\cite{eitelUncoveringConvolutionalNeural2019} was different. Instead of identifying brain regions related to the classification task, they exhibited with LRP that transfer learning between networks trained on different diseases (Alzheimer's disease to multiple sclerosis) and different MRI sequences enabled obtaining attribution maps focused on a smaller number of lesion areas. However, the authors pointed out that it would be necessary confirm their results on larger data sets.
%

Finally, Burduja et al.~\cite{burdujaAccurateEfficientIntracranial2020} trained a CNN-LSTM model to detect various hemorrhages from brain computed tomography (CT) scans. For each positive slice coming from controversial or difficult scans, they generated Grad-CAM based attribution maps and asked a group of radiologists to classify them as correct, partially correct or incorrect. This classification allowed them to determine patterns for each class of maps, and better understand which characteristics radiologists expected from these maps to be considered as correct and thus useful in practice. In particular, radiologists described maps including any type of hemorrhage as incorrect as soon as some of the hemorrhages were not highlighted, while the model only needed to detect one hemorrhage to correctly classify the slice as pathological.

\subsubsection{Comparison of several interpretability methods}

Papers described in this section used several interpretability methods and compared them in their particular context. However, as the benchmark of interpretability methods is the focus of section~\ref{subsec: which method}, which also include other types of interpretability than back-propagation, we will only focus here on what conclusions were drawn from the attribution maps.

Dyrba et al.~\cite{dyrbaComparisonCNNVisualization2020}  compared DeconvNet, guided back-propagation, deep Taylor decomposition, gradient$\odot$input, LRP (with various rules) and Grad-CAM methods for classification of Alzheimer's disease, mild cognitive impairment and normal cognition 
statuses. In accordance with the literature, they obtained a highest attention given to the hippocampus for both prodromal and demented patients.


B\"{o}hle et al.~\cite{bohleLayerwiseRelevancePropagation2019} compared two methods, LRP with $\beta$-rule and guided back-propagation for Alzheimer's disease status classification. They found that LRP attribution maps highlight the individual differences between patients, and then that they could be used as a tool for clinical guidance.

\subsection{Perturbation methods applied to neuroimaging}
\label{sec:application_peturbation}

The standard perturbation method has been widely used in the study of Alzheimer's disease \citep{baeTransferLearningPredicting2019, riekeVisualizingConvolutionalNetworks2018, nigriExplainableDeepCNNs2020, eitelTestingRobustnessAttribution2019} and related symptoms (amyloid-$\beta$ pathology) \citep{tangInterpretableClassificationAlzheimer2019}. However, most of the time, authors do not train their model with perturbed images. Hence, to generate explanation maps, the perturbation method uses images outside the distribution of the training set, which may call into question the relevance of the predictions and thus the reliability of attention maps.

\subsubsection{Variants of the perturbation method tailored to neuroimaging}

\begin{figure}[!tbh]
    \centering
    \includegraphics[width=0.7\textwidth]{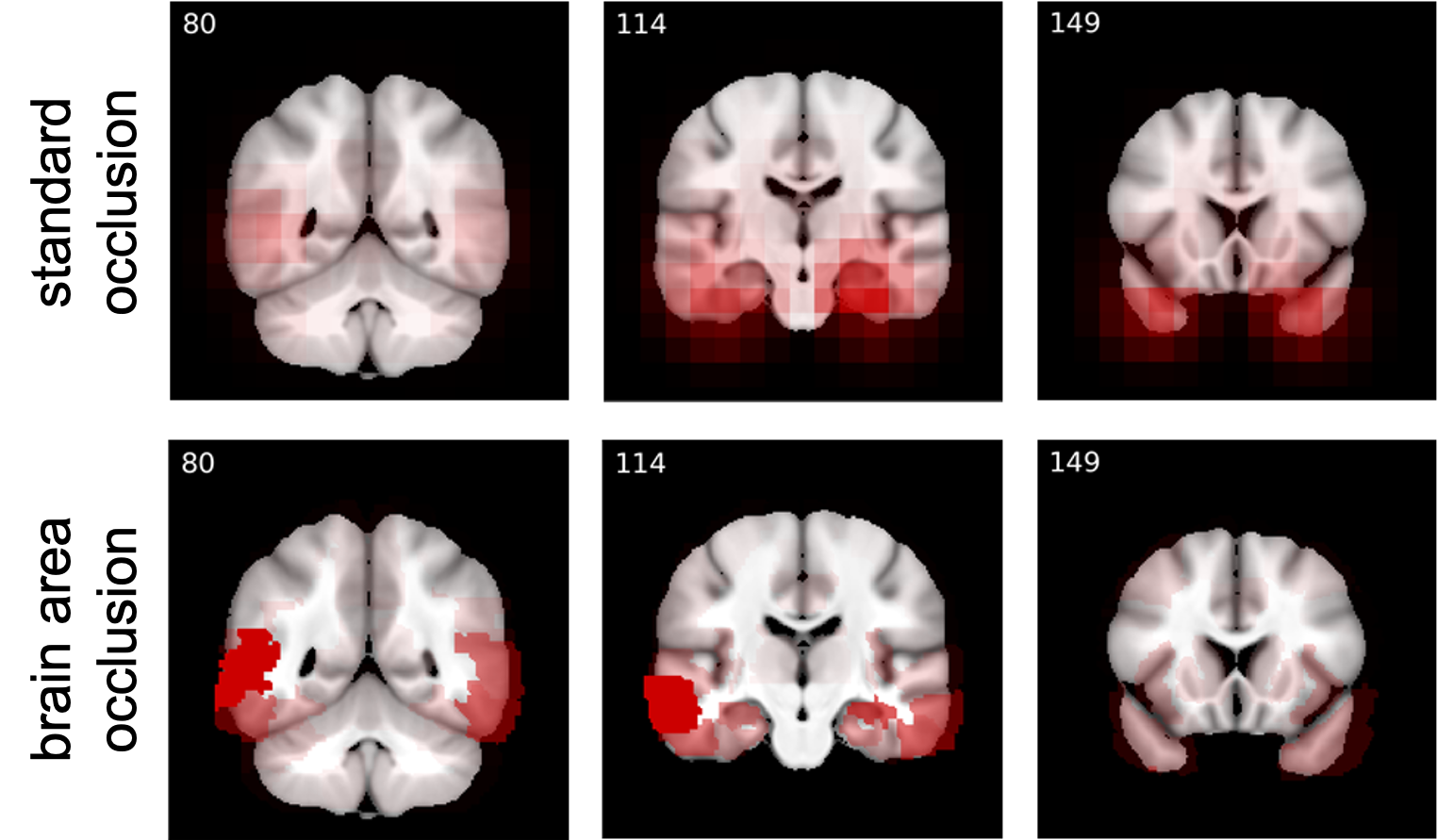}
    \caption[Mean attribution maps obtained on demented patients obtained with the standard and the brain area perturbation methods.]{Mean attribution maps obtained on demented patients. The first row corresponds to the standard and the second one to the brain area perturbation method. \\
    Reprinted by permission from Springer Nature Customer Service Centre GmbH: Springer Nature, MLCN 2018, DLF 2018, IMIMIC 2018: Understanding and Interpreting Machine Learning in Medical Image Computing Applications, \citep{riekeVisualizingConvolutionalNetworks2018}, 2018.}
    \label{fig: rieke_perturbation}
\end{figure}

Several variations of the perturbation method have been developed to adapt to neuroimaging data.
The most common variation in brain imaging is the brain area perturbation method, which consists in perturbing entire brain regions according to a given brain atlas, as done in \cite{riekeVisualizingConvolutionalNetworks2018, abrolDeepResidualLearning2020, ohClassificationVisualizationAlzheimer2019}. 
In their study of Alzheimer's disease, Abrol et al.~\cite{abrolDeepResidualLearning2020} obtained high values in their attribution maps for the usually discriminant brain regions, such as the hippocampus,the amygdala, the inferior and superior temporal gyruses, and the fusiform gyrus. Rieke et al.~\cite{riekeVisualizingConvolutionalNetworks2018} also obtained results in accordance with the medical literature, and noted that the brain area perturbation method led to a less scattered attribution map than the standard method (Figure~\ref{fig: rieke_perturbation}).
Oh et al.~\cite{ohClassificationVisualizationAlzheimer2019} used the method to compare the attribution maps of two different tasks: (1) demented patients vs cognitively normal participants and (2) stable vs progressive mild cognitively impaired patients, and noted that the regions targeted for the first task were shared with the second one (medial temporal lobe), but that some regions were specific to the second task (parts of the parietal lobe).

Guti\'{e}rrez-Becker and Wachinger~\cite{gutierrez-beckerDeepMultistructuralShape2018} adapted the standard perturbation method to a network that classified clouds of points extracted from neuroanatomical shapes of brain regions (e.g. left hippocampus) between different states of Alzheimer's disease. For the perturbation step, the authors set to $0$ the coordinates of a given point $x$ and the ones of its neighbors to then assess the relevance of the point $x$. This method allows easily generating and visualizing a 3D attribution map of the shapes under study.

\subsubsection{Advanced perturbation methods}

More advanced perturbation based methods have also been used in the literature. Nigri et al.~\cite{nigriExplainableDeepCNNs2020} compared a classical perturbation method to a swap test. The swap test replaces the classical perturbation step by a swapping step where patches are exchanged between the input brain image and a reference image chosen according to the model prediction. This exchange is possible as brain images were registered and thus brain regions are positioned in roughly the same location in each image.

\begin{figure}[!tbh]
    \centering
    \includegraphics[width=\textwidth]{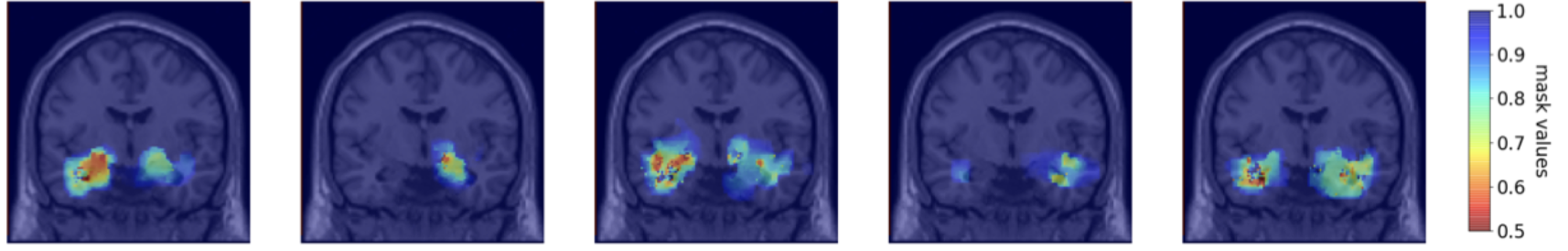}
    \caption[Attribution maps obtained with the optimized perturbation methods.]{Coronal view of the mean attribution masks on demented patients obtained for five reruns of the same network with the optimized perturbation method. \\
    Adapted with permission from Medical Imaging 2020: Image Processing, \citep{thibeau-sutreVisualizationApproachAssess2020}.}
    \label{fig: thibeausutre_occlusion}
\end{figure}

Finally, Thibeau-Sutre et al.~\cite{thibeau-sutreVisualizationApproachAssess2020} used the optimized version of the perturbation method to assess the robustness of CNNs in identifying regions of interest for Alzheimer's disease detection. They applied optimized perturbations on gray matter maps extracted from T1w MR images, and the perturbation method consisted in increasing the value of the voxels to transform patients into controls. This process aimed at simulating gray matter reconstruction to identify the most important regions that needed to be ``de-atrophied'' to be considered again as normal. However they unveiled a lack of robustness of the CNN: different retrainings led to different attribution maps (shown in Figure~\ref{fig: thibeausutre_occlusion}) even though the performance did not change.

\subsection{Distillation methods applied to neuroimaging}
\label{sec:application_distillation}

Distillation methods are less commonly used, but some very interesting use cases can be found in the literature on brain disorders, with methods such as LIME \citep{mageshExplainableMachineLearning2020} or SHAP \citep{ballIndividualVariationUnderlying2020}.

Magesh et al.~\cite{mageshExplainableMachineLearning2020} used LIME to interpret a CNN for Parkinson's disease detection from single-photon single-photon emission computed tomography (SPECT) scans. Most of the time the most relevant regions are the putamen and the caudate (which is clinically relevant), and some patients 
also showed an anomalous increase in dopamine activity in nearby areas, which is a characteristic feature of late-stage Parkinson's disease. The authors did not specify how they extracted the ``super-pixels'' necessary to the application of the method, though it could have been interesting to consider neuroanatomical regions instead of creating the voxels groups with an agnostic method.

\begin{figure}[!tbh]
    \centering
    \includegraphics[width=0.7\textwidth]{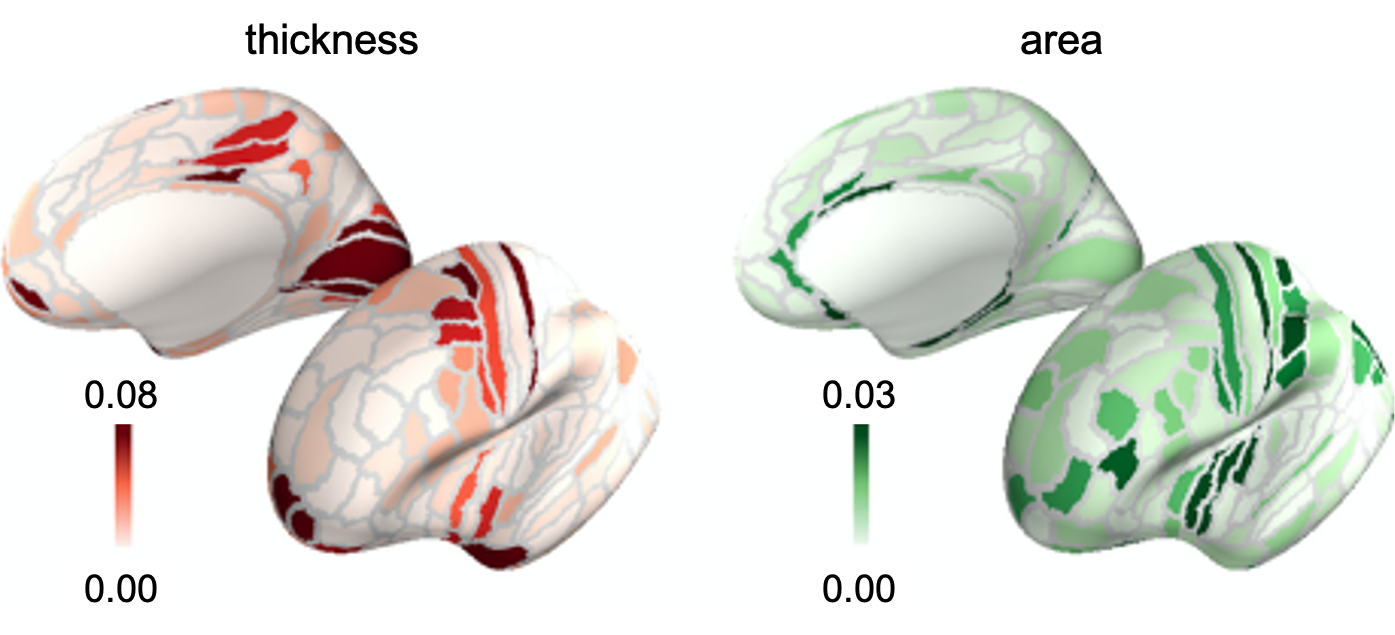}
    \caption[Mean absolute SHAP values averaged across all subjects for regional thickness and area.]{Mean absolute feature importance (SHAP values) averaged across all subjects for XGBoost on regional thicknesses (red) and areas (green).\\
    Adapted from \citep{ballIndividualVariationUnderlying2020} (CC BY 4.0).}
    \label{fig: ball_SHAP}
\end{figure}

Ball et al.~\cite{ballIndividualVariationUnderlying2020} used SHAP to obtain explanations at the individual level from three different models trained to predict participants' age from regional cortical thicknesses and areas: regularised linear model, Gaussian process regression and XGBoost,  (Figure~\ref{fig: ball_SHAP}). The authors exhibited a set of regions driving predictions for all models, and showed that regional attention was highly correlated on average with weights of the regularised linear model. However, they showed that while being consistent across models and training folds, explanations of SHAP at the individual level were generally not correlated with feature importance obtained from the weight analysis of the regularised linear model. 
The authors also exemplified that the global contribution of a region to the final prediction error (``brain age delta''), even with a high SHAP value, was in general small, which indicated that this error was best explained by changes spread across several regions~\cite{ballIndividualVariationUnderlying2020}. 

\subsection{Intrinsic methods applied to neuroimaging}
\label{sec:application_intrinsic}

\subsubsection{Attention modules}

Attention modules have been increasingly used in the past couple of years, as they often allow a boost in performance while being rather easy to implement and interpret. 
To diagnose various brain diseases from brain CT images,
Fu et al.~\cite{fuAttentionbasedFullSlice2021} built a model integrating a ``two step attention'' mechanism that selects both the most important slices and the most important pixels in each slice. The authors then leveraged these attention modules to retrieve the five most suspicious slices and highlight the areas with the more significant attention. 

\begin{figure}[!tbh]
    \centering
    \includegraphics[width=0.9\textwidth]{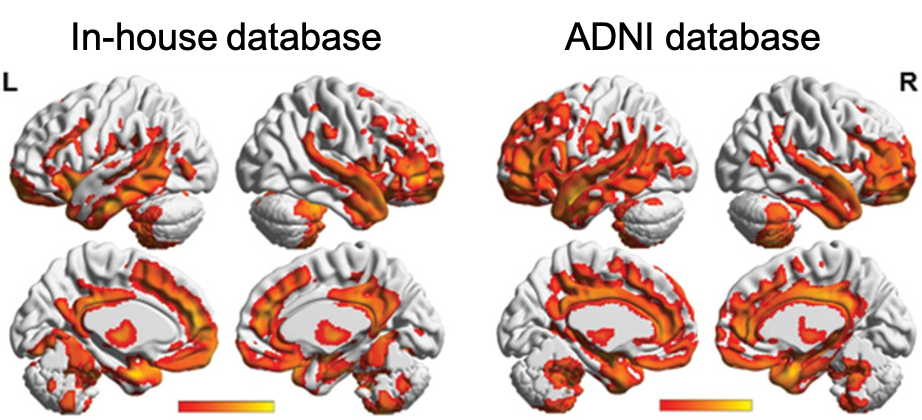}
    \caption[Attribution maps generated by an attention mechanism module.]{Attribution maps (left: in-house database, right: ADNI database) generated by an attention mechanism module, indicating the discriminant power of various brain regions for Alzheimer's disease diagnosis. \\
    Adapted from \citep{jinGeneralizableReproducibleNeuroscientifically2020} (CC BY 4.0). }
    \label{fig: jin_attention}
\end{figure}

In their study of Alzheimer's disease,  
Jin et al.~\cite{jinGeneralizableReproducibleNeuroscientifically2020} used a 3D attention module to capture the most discriminant brain regions used for Alzheimer's disease diagnosis. As shown in Figure~\ref{fig: jin_attention}, they obtained significant correlations between attention patterns for two independent databases. They also obtained significant correlations between regional attention scores of two different databases, which indicated a strong reproducibility of the results.

\subsubsection{Modular transparency}

Modular transparency has often been used in brain imaging analysis. A possible practice consists in first generating a target probability map of a black-box model, before feeding this map to a classifier to generate a final prediction, as done in \citep{qiuDevelopmentValidationInterpretable2020, leeInterpretableAlzheimerDisease2019}. 

\begin{figure}[!tbh]
    \centering
    \includegraphics[width=0.9\textwidth]{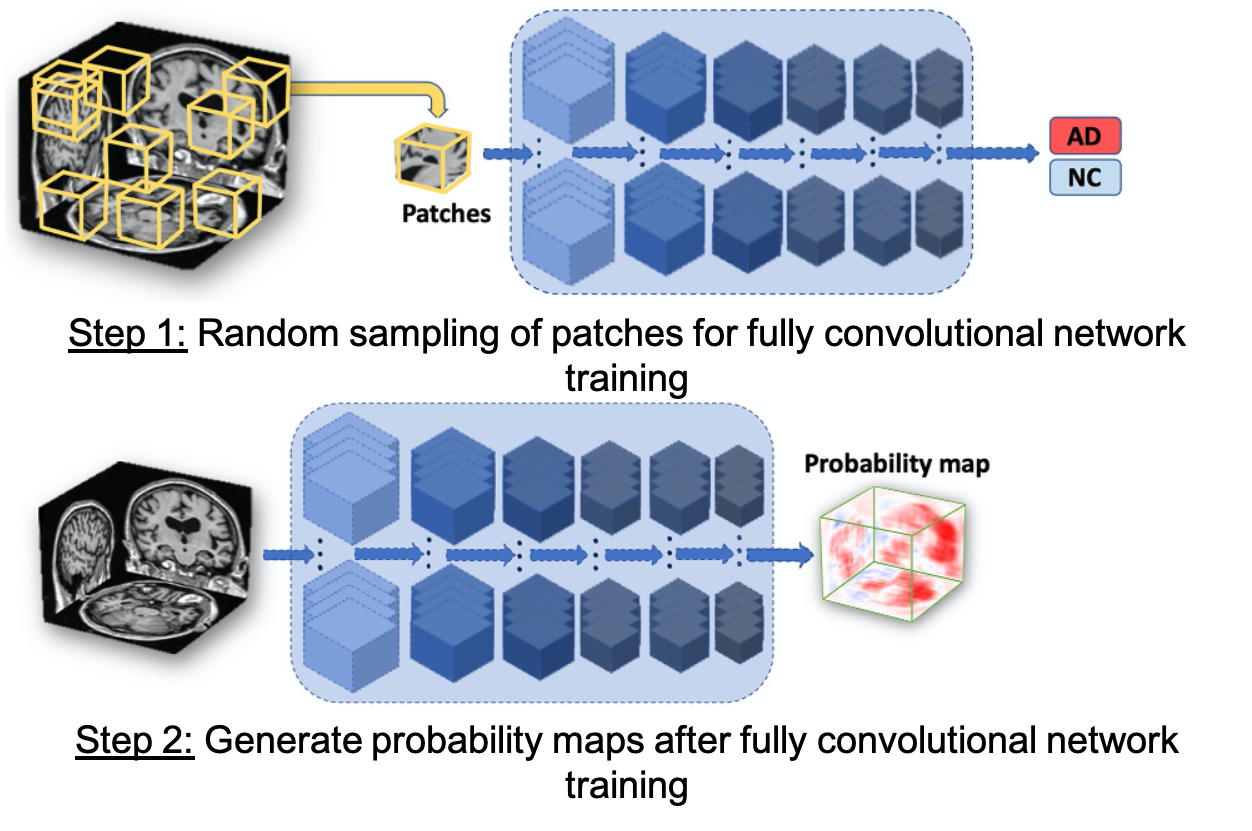}
    \caption[Example of modular transparency using random patch learning.]{Randomly selected samples of T1-weighted full MRI volumes are used as input to learn the Alzheimer's disease status at the individual level (Step 1). The application of the model to whole images leads to the generation of participant-specific disease probability maps of the brain (Step 2). \\
    Adapted from Brain: A Journal of Neurology, 143, \citep{qiuDevelopmentValidationInterpretable2020}, 2020, with permission of Oxford University Press. }
    \label{fig: qiu_modular}
\end{figure}

Qiu et al.~\cite{qiuDevelopmentValidationInterpretable2020} used a convolutional network to generate an attribution map from patches of the brain, highlighting brain regions associated with Alzheimer's disease diagnosis (see Figure~\ref{fig: qiu_modular}).
Lee et al.~\cite{leeInterpretableAlzheimerDisease2019} first parcellated gray matter density maps into 93 regions. For each of these regions, several deep neural networks were trained on randomly selected voxels and their outputs were averaged to obtain a mean regional disease probability. Then, by concatenating these regional probabilities, they generated a region-wise disease probability map of the brain, which was further used to perform Alzheimer's disease detection.

\begin{figure}[!tbh]
    \centering
    \includegraphics[width=0.9\textwidth]{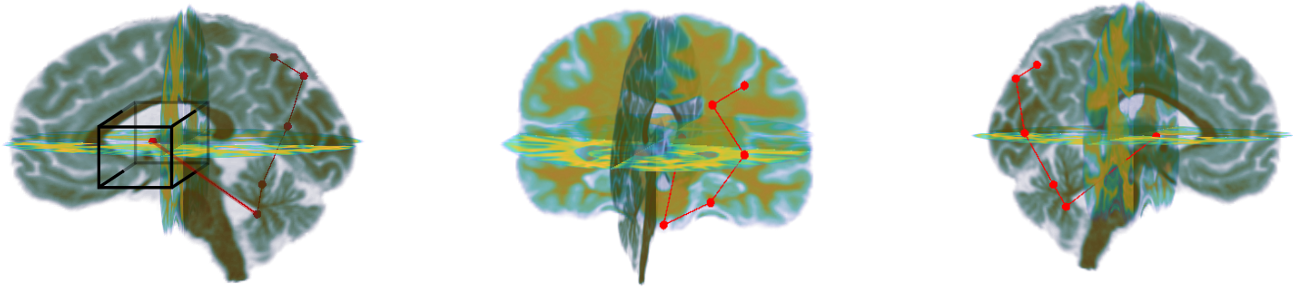}
    \caption[Trajectory taken by the framework trained based on the work of Ba et al.~\citep{baMultipleObjectRecognition2015}.]{Trajectory taken by the framework for a participant from the ADNI test set. A bounding box around the first location attended to is included to indicate the approximate size of the glimpse that the recurrent neural network receives; this is the same for all subsequent locations. \\
    Adapted from \citep{woodNEURODRAM3DRecurrent2019}. Permission to reuse was kindly granted by the authors.}
    \label{fig: wood_modular}
\end{figure}

The approach of Ba et al.~\cite{baMultipleObjectRecognition2015} was also applied to Alzheimer's disease detection~\cite{woodNEURODRAM3DRecurrent2019} (preprint). Though that work is still a preprint, the idea is interesting as it aims at reproducing the way a radiologist looks at an MR image. The main difference with \cite{baMultipleObjectRecognition2015} is the initialization, as the context network does not take as input the whole image but clinical data of the participant. Then the framework browses the image in the same way as in the original paper: a patch is processed by a recurrent neural network and from its internal state the glimpse network learns which patch should be looked at next. After a fixed number of iterations, the internal state of the recurrent neural network is processed by a classification network that gives the final outcome. The whole system is interpretable as the trajectory of the locations (illustrated in Figure~\ref{fig: wood_modular}) processed by the framework allows understanding which regions are more important for the diagnosis. However this framework may have a high dependency to clinical data: as the initialization depends on scores used to diagnose Alzheimer's disease, the classification network may learn to classify based on the initialization only and most of the trajectory may be negligible to assess the correct label.

\begin{figure}[!tbh]
    \centering
    \includegraphics[width=0.8\textwidth]{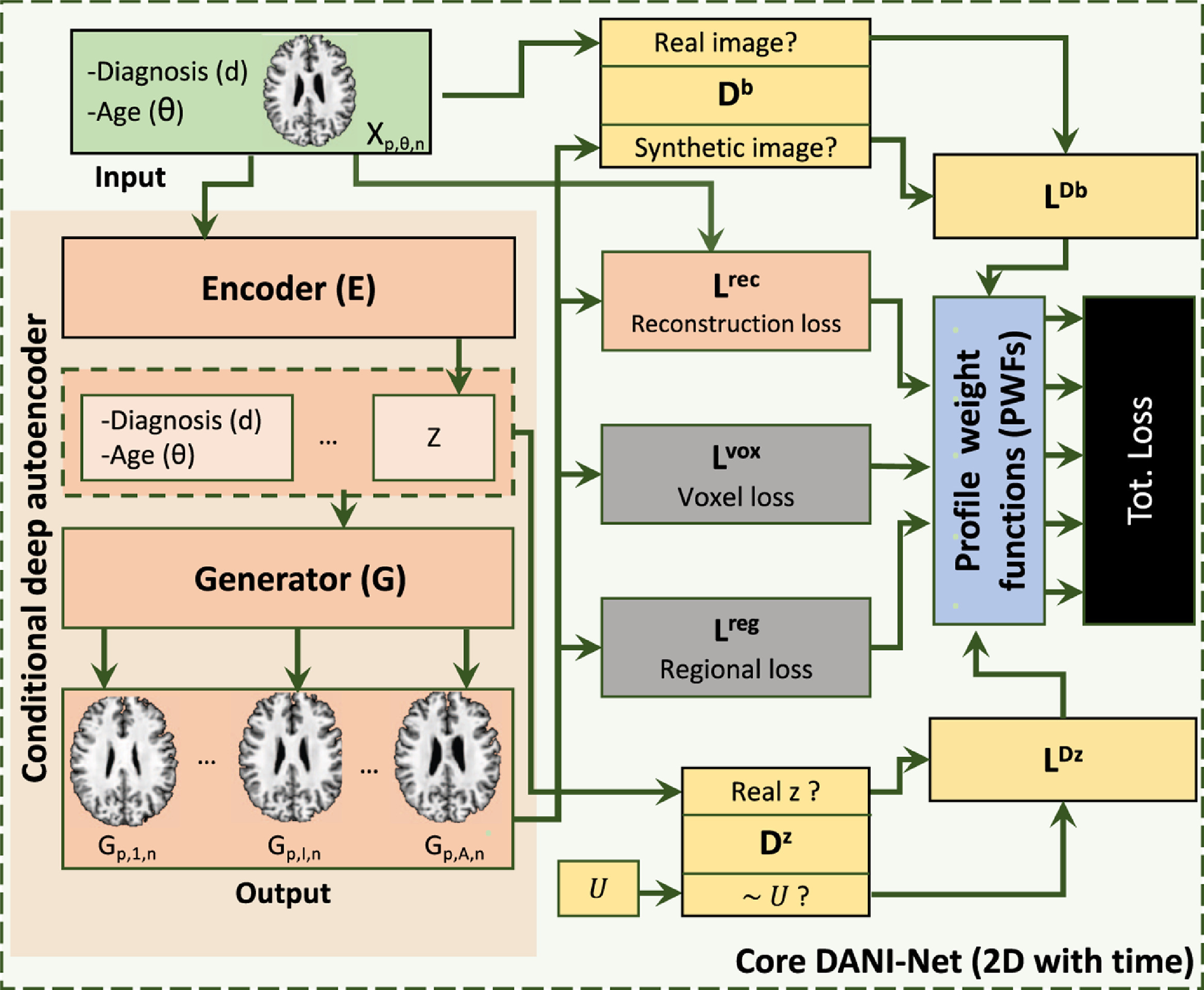}
    \caption[Pipeline used for training the DaniNet framework.]{Pipeline used for training the proposed DaniNet framework that aims to learn a longitudinal model of the progression of Alzheimer's disease. \\
    Adapted from \citep{raviDegenerativeAdversarialNeuroimage2022} (CC BY 4.0).}
    \label{fig: ravi_modular}
\end{figure}

Another framework, the DaniNet, proposed by Ravi et al.~\cite{raviDegenerativeAdversarialNeuroimage2022}, is composed of multiple networks, each with a defined function, as illustrated in Figure~\ref{fig: ravi_modular}.
\begin{itemize}
    \item The conditional deep autoencoder (in orange) learns to reduce the size of the slice $x$ to a latent variable $Z$ (encoder part), and then to reconstruct the original image based on $Z$ and two additional variables: the diagnosis and age (generator part). Its performance is evaluated thanks to the reconstruction loss $L^{rec}$.
    \item Discriminator networks (in yellow) either force the encoder to take temporal progression into account ($D_z$) or try to determine if the output of the generator are real or generated images ($D_b$).
    \item Biological constraints (in grey) force the previous generated image of the same participant to be less atrophied than the next one (voxel loss) and learn to find the diagnosis thanks to regions of the generated images (regional loss).
    \item Profile weight functions (in blue) aim at funding appropriate weights for each loss to compute the total loss.
\end{itemize}
The assembly of all these components allows learning a longitudinal model that characterizes the progression of the atrophy of each region of the brain. This atrophy evolution can then be visualized thanks to a neurodegeneration simulation generated by the trained model by sampling missing intermediate values.

\subsection{Benchmarks conducted in the literature}
\label{subsec:benchmarks}

This section describes studies that compared several interpretability methods. We separated evaluations based on metrics from those which are purely qualitative. Indeed, even if the interpretability metrics are not mature yet, it is essential to try to measure quantitatively the difference between methods rather than to only rely on human perception, which may be biased.

\subsubsection{Quantitative evaluations}

Eitel and Ritter~\cite{eitelTestingRobustnessAttribution2019} tested the robustness of four methods: standard perturbation, gradient$\odot$input, guided back-propagation and LRP. To evaluate these methods, the authors trained ten times the same model with a random initialization and generated attribution maps for each of the ten runs. For each method, they exhibited significant differences between the averaged true positives/negatives attribution maps of the ten runs. To quantify this variance, they computed the $\ell$2-norm between the attribution maps, and determined for each model the brain regions with the highest attribution. They concluded that LRP and guided back-propagation were the most consistent methods, both in terms of distance between attribution maps and most relevant brain regions. However this study makes a strong assumption: to draw these conclusions, the network should provide stable interpretations across retrainings. Unfortunately, Thibeau-Sutre et al.~\cite{thibeau-sutreVisualizationApproachAssess2020} showed that the study of the robustness of the interpretability method and of the network should be done separately, as their network retraining was not robust. Indeed, they first showed that the interpretability method they chose (optimized perturbation) was robust according to different criteria, then they observed that network retraining led to different attribution maps. The robustness of an interpretability method thus cannot be assessed from the protocol described in~\cite{eitelTestingRobustnessAttribution2019}. Moreover, the fact that guided back-propagation is one of the most stable method meets the results of \cite{adebayoSanityChecksSaliency2018}, who observed that guided back-propagation always gave the same result independently from the weights learned by a network (see Section~\ref{sec:theoretical_limitations}). 

B\"{o}hle et al.~\cite{bohleLayerwiseRelevancePropagation2019} measured the benefit of LRP with $\beta$-rule compared to guided back-propagation by comparing the intensities of the mean attribution map of demented patients and the one of cognitively normal controls. They concluded that LRP allowed a stronger distinction between these two classes than guided back-propagation, as there was a greater difference between the mean maps for LRP. Moreover, they found a stronger correlation between the intensities of the LRP attribution map in the hippocampus and the hippocampal volume than for guided back-propagation. But as \cite{adebayoSanityChecksSaliency2018} demonstrated that guided back-propagation has serious flaws, it does not allow drawing strong conclusions.

Nigri et al.~\cite{nigriExplainableDeepCNNs2020} compared the standard perturbation method to a swap test (see Section~\ref{sec:application_peturbation}) using two properties: the continuity and the sensitivity. The continuity property is verified if two similar input images have similar explanations. The sensitivity property affirms that the most salient areas in an explanation map should have the greater impact in the prediction when removed. The authors carried out experiments with several types of models, and both properties were consistently verified for the swap test, while the standard perturbation method showed a significant absence of continuity and no conclusive fidelity values~\cite{nigriExplainableDeepCNNs2020}.

Finally Rieke et al.~\cite{riekeVisualizingConvolutionalNetworks2018} compared four visualization methods: standard back-propagation, guided back-propagation, standard perturbation and brain area perturbation. They computed the Euclidean distance between the mean attribution maps of the same class for two different methods and observed that both gradient methods were close, whereas brain area perturbation was different from all others. They concluded that as interpretability methods lead to different attribution maps, one should compare the results of available methods and not trust only one attribution map.

\subsubsection{Qualitative evaluations}

Some works compared interpretability methods using a purely qualitative evaluation.

First, Eitel et al.~\cite{eitelUncoveringConvolutionalNeural2019} generated attribution maps using the LRP and gradient$\odot$input methods and obtained very similar results. This could be expected as it was shown that there is a strong link between LRP and gradient$\odot$input (see Section~\ref{subsubsec: Relevance BP}). 

Dyrba et al.~\cite{dyrbaComparisonCNNVisualization2020} compared DeconvNet, guided back-propagation, deep Taylor decomposition, gradient$\odot$input, LRP (with various rules) and Grad-CAM. The different methods roughly exhibited the same highlighted regions, but with a significant variability in focus, scatter and smoothness, especially for the Grad-CAM method. These conclusions were derived from a visual analysis. According to the authors, LRP and deep Taylor decomposition delivered the most promising results with a highest focus and less scatter~\cite{dyrbaComparisonCNNVisualization2020}.

Tang et al.~\cite{tangInterpretableClassificationAlzheimer2019} compared two interpretability methods that seemed to have different properties: guided Grad-CAM would provide a fine-grained view of feature salience, whereas standard perturbation highlights the interplay of features among classes. A similar conclusion was drawn by Rieke et al.~\cite{riekeVisualizingConvolutionalNetworks2018}.

\subsubsection{Conclusions from the benchmarks}

The most extensively compared method is LRP, and each time it has been shown to be the best method compared to others. However, its equivalence with gradient$\odot$input for networks using ReLU activations still questions the usefulness of the method, as gradient$\odot$input is much easier to implement. Moreover, the studies reaching this conclusion are not very insightful: \cite{eitelTestingRobustnessAttribution2019} may suffer from methodological biases, \cite{bohleLayerwiseRelevancePropagation2019} compared LRP only to guided back-propagation, which was shown to be irrelevant \citep{adebayoSanityChecksSaliency2018}, and \cite{dyrbaComparisonCNNVisualization2020} only performed a qualitative assessment.

As proposed in conclusion by Rieke et al.~\cite{riekeVisualizingConvolutionalNetworks2018}, a good way to assess the quality of interpretability methods could be to produce some form of ground truth for the attribution maps, for example by implementing simulation models that control for the level of separability or location of differences.

\section{Limitations and recommendations}
\label{sec:limitations}

Many methods have been proposed for interpretation of deep learning models. The field is not mature yet and none of them has become a standard. Moreover, a large panel of studies have been applied to neuroimaging data, but the value of the results obtained from the interpretability methods is often still not clear. Furthermore, many applications suffer from methodological issues, making their results (partly) irrelevant. In spite of this, we believe that using interpretability methods is highly useful, in particular to spot cases where the model exploits biases in the dataset.

\subsection{Limitations of the methods}
\label{sec:theoretical_limitations}

It is not often clear whether the interpretability methods really highlight features relevant to the algorithm they interpret. This way, Adebayo et al.~\cite{adebayoSanityChecksSaliency2018} showed that the attribution maps produced by some interpretability methods (guided back-propagation and guided Grad-CAM) may not be correlated at all with the weights learned by the network during its training procedure. They prove it with a simple test called ``cascading randomization''. In this test, the weights of a network trained on natural images are randomized layer per layer, until the network is fully randomized. At each step, they produce an attribution map with a set of interpretability methods to compare it to the original ones (attribution maps produced without randomization). In the case of guided back-propagation and guided Grad-CAM, all attribution maps were identical, which means that the results of these methods were independent of the training procedure.

Unfortunately, this type of failures does not only affect interpretability methods but also the metrics designed to evaluate their reliability, which makes the problem even more complex. Tomsett et al.~\cite{tomsettSanityChecksSaliency2020} investigated this issue by evaluating interpretability metrics with three properties:
\begin{itemize}
    \item \textbf{inter-rater interpretability} assesses whether a metric always rank different interpretability methods in the same way for different samples in the data set,
    \item \textbf{inter-method reliability} checks that the scores given by a metric on each saliency method fluctuate in the same way between images, 
    \item \textbf{internal consistency} evaluates if different metrics measuring the same property (for example fidelity) produce correlated scores on a set of attribution maps.
\end{itemize}
They concluded that the investigated metrics were not reliable, though it is difficult to know the origin of this unreliability due to the tight coupling of model, interpretability method and metric. 


\subsection{Methodological advice}
\label{sec:methodological_advice}
Using interpretability methods is more and more common in medical research. Even though this field is not yet mature and the methods have limitations, we believe that using an interpretability method is usually a good thing because it may spot cases where the model took decisions from irrelevant features. However, there are methodological pitfalls to avoid and good practices to adopt to make a fair and sound analysis of your results.

You should first clearly state in your paper which interpretability method you use as there exist several variants for most of the methods (see section~\ref{sec:section2}), and its parameters should be clearly specified. Implementation details may also be important: for the Grad-CAM method, attribution maps can be computed at various levels in the network; for a perturbation method, the size and the nature of the perturbation greatly influence the result.
The data on which methods are applied should also be made explicit: for a classification task, results may be completely different if samples are true positives or true negatives, or if they are taken from the train or test sets.

Taking a step back from the interpretability method and especially attribution maps is fundamental as they present several limitations~\cite{bohleLayerwiseRelevancePropagation2019}.
First, there is no ground truth for such maps, which are usually visually assessed by authors. Comparing obtained results with the machine learning literature is a good first step, but be aware that you will most of the time find a paper to support your findings, so we suggest to look at established clinical references.
Second, attribution maps are usually sensitive to the interpretability method, its parameters (e.g. $\beta$ for LRP), but also to the final scale used to display maps. A slight change in one of these variables may significantly impact the interpretation.
Third, an attribution map is a way to measure the impact of pixels on the prediction of a given model, but it does not provide underlying reasons (e.g. pathological shape) or explain potential interactions between pixels. A given pixel might have a low attribution when considered on its own, but have a huge impact on the prediction when combined with another.
Fourth, the quality of a map strongly depends on the performance of the associated model. Indeed, low performance models are more likely to use wrong features. However, even in this case, attribution maps may be leveraged, e.g. to determine if the model effectively relies on irrelevant features (such as visual artefacts) or if there are biases in the data set~\cite{lapuschkinAnalyzingClassifiersFisher2016}.

One must also be very careful when trying to establish new medical findings using model interpretations, as we do not always know how the interpretability methods react when applied to correlated features. Then even if a feature seems to have no interest for a model, this does not mean that it is not useful in the study of the disease (for example, a model may not use information from the frontal lobe when diagnosing Alzheimer's disease dementia, but this does not mean that this region is not affected by the disease).

Finally, we suggest implementing different interpretability methods to obtain complementary insights from attribution maps. For instance, using LRP in addition to the standard back-propagation method provides a different type of information, as standard back-propagation gives the sensibility of the output with respect to the input, while LRP shows the contribution of each input feature to the output. Moreover, using several metrics allows a quantitative comparison between them using interpretability metrics (see section~\ref{sec:evaluation_metrics}).


\subsection{Which method should I choose?}
\label{subsec: which method}

We conclude this section on how to choose an interpretability method. Some benchmarks were conducted to assess the properties of some interpretability methods compared to others (see Section~\ref{subsec:benchmarks}). Though these are good initiatives, there are still not enough studies (and some of them suffer from methodological flaws) to draw solid conclusions. This is why we give in this section some practical advice to the reader to choose an interpretability method based on more general concepts.

Before implementing an interpretability method, we suggest reviewing the following points to help you choose carefully.
\begin{itemize}
    \item \textbf{Implementation complexity}\quad Some methods are more difficult to implement than others, and may require substantial coding efforts. However, many of them have already been implemented in libraries or github repositories (e.g. \cite{uozbulak_pytorch_vis_2021}), so we suggest looking online before trying to re-implement them. This is especially true for model-agnostic methods, such as LIME, SHAP or perturbations, for which no modification of your model is required. For model-specific methods, such as back-propagation ones, the implementation will depend on the model, but if its structure is a common one (e.g. regular CNN with feature extraction followed by a classifier), it is also very likely that an adequate implementation is already available  (e.g. Grad-CAM on CNN in \cite{uozbulak_pytorch_vis_2021}).
    \item \textbf{Time cost}\quad Computation time greatly differs from one method to another, especially when input data is heavy. For instance, perturbing high dimension images is time expensive, and it would be much faster to use standard back-propagation.
    \item \textbf{Method parameters}\quad The number of parameters to set varies between methods, and their choice may greatly influence the result. For instance, the patch size, the step size (distance between two patches) as well as the type of perturbation (e.g. white patches or blurry patches) must be chosen for the standard perturbation method, while the standard back-propagation does not need any parameter. Thus, without prior knowledge on the interpretability results, methods with no or only a few parameters are a good option.
    \item \textbf{Literature}\quad Finally, our last piece of advice is to look into the literature to determine the methods that have commonly been used in  your domain of study. A highly used method does not guarantee its quality (e.g. guided back-propagation~\cite{adebayoSanityChecksSaliency2018}), but it is usually a good first try.
\end{itemize}
To sum up, we suggest that you choose (or at least begin with) an interpretability method that is easy to implement, time efficient, with no parameters (or only a few) to tune and commonly used. In the context of brain image analysis, we suggest using the standard back-propagation or Grad-CAM methods.
Before using a method you do not know well, you should check that other studies did not show that this method is not relevant (which is the case for guided back-propagation or guided Grad-CAM), or that it is not equivalent to another method (for example LRP on networks with ReLU activation layers and gradient$\odot$input). 

Regarding interpretability metrics, there is no consensus in the community as the field is not mature yet. General advice would be to use different metrics and confront them to human observers, taking for example the methodology described in~\cite{ribeiroWhyShouldTrust2016}.

\section{Conclusion}

Interpretability of machine learning models is an important topic, in particular in the medical field. First, this is a natural need expressed by clinicians who are potential users of medical decision support systems. Moreover, it has been shown in many occasions that models with high performance can actually be using irrelevant features. This is dangerous because it means that they are exploiting biases in the training data sets and thus may dramatically fail when applied to new data sets or deployed in clinical routine.

Interpretability is a very active field of research and many approaches have been proposed. They have been extensively applied in neuroimaging, and very often allowed highlighting clinically relevant regions of the brain that were used by the model. However, comparative benchmarks are not entirely conclusive and it is currently not clear which approach is the most adapted for a given aim. In other words, it is very important to keep in mind that the field of interpretability is not yet mature. It is not yet clear which are the best methods or even if the most widely used approaches will still be considered a standard in the near future. 

That being said, we still strongly recommend that a classification or regression model be studied with at least one interpretability method. Indeed, evaluating the performance of the model is not sufficient in itself and the additional use of an interpretation method may allow detecting biases and models that perform well but for bad reasons and thus would not generalize to other settings.

\clearpage

\section*{Appendices}
\renewcommand{\thesubsection}{\Alph{subsection}}
\setcounter{subsection}{0}

\subsection{Short reminder on network training procedure}
\label{appendix:network}

During the training phase, a neural network updates its weights to make a series of inputs match with their corresponding target labels:
\begin{enumerate}
    \item \textit{Forward pass}\quad The network processes the input image to compute the output value.
    \item \textit{Loss computation}\quad The difference between the true labels and the output values is computed according to a criterion (cross-entropy, mean squared error...). This difference is called the loss, and should be as low as possible
    \item \textit{Backward pass}\quad For each learnable parameter of the network, the gradients with respect to the loss are computed.
    \item \textit{Weight update}\quad Weights are updated according to the gradients and an optimizer rule (stochastic gradient descent, Adam, Adadelta...).
\end{enumerate}
As a network is a composition of functions, the gradients of the weights of a layer $l$ with respect to the loss can be easily obtained according to the values of the gradients in the following layers. This way of computing gradients layer per layer is called back-propagation.

\subsection{Description of the main brain disorders mentioned in the reviewed studies}
\label{appendix:diseases}

This appendix aims at shortly presenting the diseases considered by the studies reviewed in Section~\ref{sec:section3}.

The majority of the studies focused on the classification of Alzheimer's disease (AD), a neurodegenerative disease of the elderly. Its pathological hallmarks are senile plaques formed by amyloid-$\beta$ protein and neurofibrillary tangles that are tau protein aggregates. Both can be measured in vivo using either PET imaging or CSF biomarkers. Several other biomarkers of the disease exist. In particular, atrophy of gray and white matter measured from T1w MRI is often used, even though it is not specific of AD.
There is strong and early atrophy in the hippocampi that can be linked to the memory loss, even though other clinical signs are found and other brain areas are altered.
The following diagnosis statuses are often used:
\begin{itemize}
    \item \textbf{AD} refers to demented patients,
    \item \textbf{CN} refers to cognitively normal participants,
    \item \textbf{MCI} refers to patients in with mild cognitive impairment (they have an objective cognitive decline but it is not sufficient yet to cause a loss of autonomy),
    \item \textbf{stable MCI} refers to MCI patients who stayed stable during a defined period (often three years),
    \item \textbf{progressive MCI} refers to MCI patients who progressed to Alzheimer's disease during a defined period (often three years).
\end{itemize}
Most of the studies analysed T1w MRI data, except \cite{tangInterpretableClassificationAlzheimer2019} where the patterns of amyloid-$\beta$ in the brain are studied.

Fronto-temporal dementia is another neurodegenerative disease in which the neuronal loss dominates in the frontal and temporal lobes. Behavior and language are the most affected cognitive functions.

Parkinson's disease is also a neurodegenerative disease. It primarily affects dopaminergic neurons in the substantia nigra. A commonly used neuroimaging technique to detect this loss of dopaminergic neurons is the SPECT, as it uses a ligand that binds to dopamine transporters. Patients are affected by different symptoms linked to motor faculties such as tremor, slowed movements and gait disorder, but also sleep disorder, depression and other symptoms.

Multiple sclerosis is a demyelinating disease with a neurodegenerative component affecting younger people (it begins between the ages of 20 and 50). It causes demyelination of the white matter in the brain (brain stem, basal ganglia, tracts near the ventricles), optic nerve and spinal cord. This demyelination results in autonomic, visual, motor and sensory problems.

Intracranial hemorrhage may result from a physical trauma or nontraumatic causes such as a ruptured aneurysm. Different subtypes exist depending on the location of the hemorrhage. 

Autism is a spectrum of neurodevelopmental disorders affecting social interation and communication. Diagnosis is done based on clinical signs (behavior) and the patterns that may exist in the brain are not yet reliably described as they overlap with the neurotypical population.

Some brain characteristics that may be related to brain disorders and detected in CT scans were considered in the data set CQ500:
\begin{itemize}
    \item \textbf{Midline Shift} is a shift of the center of the brain past the center of the skull.
    \item \textbf{Mass Effect} is caused by the presence of an intracranial lesion (for example a tumor) that is compressing nearby tissues.
    \item \textbf{Calvarial Fractures} are fractures of the skull.
\end{itemize}

Finally, one study \citep{ballIndividualVariationUnderlying2020} learned to predict the age of cognitively normal patients. Such algorithm can help in diagnosing brain disorders as patients will have a greater brain age than their chronological age, then it establishes that a participant is not in the normal distribution.

\clearpage

\section*{Acknowledgments}

The research leading to these results has received funding from the French government under management of Agence Nationale de la Recherche as part of the ``Investissements d'avenir'' program, reference ANR-19-P3IA-0001 (PRAIRIE 3IA Institute) and reference ANR-10-IAIHU-06 (Agence Nationale de la Recherche-10-IA Institut Hospitalo-Universitaire-6).

\clearpage
\bibliographystyle{spbasicsort}
\bibliography{references}

\end{document}

%% file: template.tex
\usepackage[noblocks]{authblk} 
\usepackage[numbers]{natbib}
\usepackage{graphicx}
\usepackage{fancyhdr}
\usepackage[colorlinks=true, linkcolor=MidnightBlue, urlcolor=MidnightBlue, citecolor=PineGreen]{hyperref}
\usepackage[tableposition=top, justification=justified, textfont=it]{caption}
\usepackage{titlesec}
\usepackage{multicol}
\usepackage{afterpage}
\usepackage{lipsum}
\usepackage[dvipsnames, x11names]{xcolor}
\usepackage[inner=4cm, outer=4cm, top=2.8cm, bottom=2.8cm, marginparsep=0cm, marginparwidth=0cm, includehead, includefoot]{geometry}
\usepackage{setspace}
\usepackage{array}
\usepackage{bm}
\usepackage{tikz}
\usepackage{float}
\usepackage[framemethod]{mdframed}
\usepackage{amsmath}
\usepackage{amssymb}
\usepackage{rotating}
\usepackage{tikz}
\usetikzlibrary{trees,positioning,shapes,shadows,arrows}
\usepackage{enumitem}
\setlist{nosep}
\usepackage{natbib}
\usepackage{amsmath}
\usepackage{amssymb} 
\usepackage{graphicx}
\usepackage{float}
\usepackage{url}
\usepackage{fancyhdr}
\usepackage{hyperref}
\usepackage{bbold}
\usepackage{caption}
\usepackage{pdflscape}
\usepackage{subcaption}
\usepackage{latexsym,pictex}
\usepackage{amsfonts}
\usepackage{setspace}
\usepackage{makecell}

\usepackage{color}
\usepackage[dvipsnames]{xcolor}
\usepackage{multirow}

\tikzstyle{root}=[rectangle, draw=black, rounded corners, fill=lightgray, drop shadow, text centered, anchor=north, text=black, text width=5cm, font=\small]
\tikzstyle{category}=[rectangle, draw=black, rounded corners, fill=lightgray, drop shadow, text centered, anchor=north, text=black, text width=3.5cm, font=\footnotesize]
\tikzstyle{description}=[rectangle, draw=black, rounded corners, fill=white, drop shadow, anchor=north, text=black, text width=3.5cm, font=\scriptsize]
\tikzstyle{myarrow}=[stealth-, thick]

\titleformat{\section}{\Large \bfseries \centering \scshape}{\thesection.}{0.3em}{}[{\titlerule[0.5pt]}]

\definecolor{shadecolor}{RGB}{230,230,230}
\newcommand{\mybox}[1]{\par\noindent\colorbox{shadecolor}
{\parbox{\dimexpr\textwidth-2\fboxsep\relax}{#1}}}

\titleformat{\subsection}{\large \bfseries \mybox}{\thesubsection}{1em}{}

\titleformat{\subsubsection}{\itshape}{\thesubsubsection.}{0.3em}{}

\renewenvironment{abstract}
{\vskip 2.5ex {\large\bf\noindent Abstract}\vspace{0.7ex} \\ %
  \bgroup\noindent\ignorespaces}%
{\par\egroup\vskip 2.5ex}

\newenvironment{keywords}
{\bgroup\leftskip 20pt\rightskip 20pt \small\noindent{\bf Keywords:} }%
{\par\egroup\vskip 10ex}

\fancypagestyle{firstpage}{%
  
  \fancyhead[]{}
  \fancyfoot[C]{}
  \fancyfoot[L]{\footnotesize To appear in \\
  O. Colliot (Ed.), \textit{Machine Learning for Brain Disorders}, Springer\\
  }

}

\fancypagestyle{otherpages}{%
  
  \fancyhead[LE]{\thepage}
  \fancyhead[RE]{\runningauthor}
  \fancyhead[LO]{\runningheadtitle}
  \fancyhead[RO]{\thepage}
  \fancyfoot[L]{\footnotesize  \textit{Machine Learning for Brain Disorders}, Chapter~\chapternumber}
  \fancyfoot[C]{}
}

\makeatletter
\renewcommand{\maketitle}{\bgroup\setlength{\parindent}{0pt}

\begin{flushright}
  \color{MidnightBlue}
  \textbf{\LARGE Chapter~\chapternumber}
\end{flushright}

\vspace{0.3in}

\begin{flushleft}
    \setstretch{2.0} 
    \textbf{\color{MidnightBlue}\huge\@title}
\end{flushleft}

\vspace{0.15in}

\begin{flushleft}
    \textbf{\bfseries \large\@author}
\end{flushleft}\egroup
}

\makeatother

\DeclareCaptionLabelFormat{labelformat}{\color{MidnightBlue}\bfseries #1 #2}
\renewcommand{\bibpreamble}{\scriptsize \begin{multicols}{2}}
\renewcommand{\bibpostamble}{\end{multicols}}

\mdfsetup{skipabove=\topskip, skipbelow=\topskip}

\newcounter{nicebox}

\newfloat{floatbox}{thp}{lop}
\floatname{floatbox}{Box}

\DeclareMathAlphabet{\mathsfit}{\encodingdefault}{\sfdefault}{m}{sl}
\SetMathAlphabet{\mathsfit}{bold}{\encodingdefault}{\sfdefault}{bx}{n}

%% file: studies_table_wo_preprocessing.tex
\afterpage{%
\clearpage
\thispagestyle{empty}
\begin{landscape}

\begin{table}[!t]
    \centering
    \caption[Summary of the studies applying interpretability methods to neuroimaging data]{Summary of the studies applying interpretability methods to neuroimaging data which are presented in Section~\ref{sec:section3}.}
    \label{tab:section3}

    \begin{tabular}{|m{0.35\textwidth}<{\centering}|m{0.25\textwidth}<{\centering}|m{0.15\textwidth}<{\centering}|m{0.3\textwidth}<{\centering}|m{0.4\textwidth}<{\centering}|m{0.1\textwidth}<{\centering}|}
    \hline
    Study & Data set & Modality & Task & Interpretability method & Section \\
    \hline
    \hline
    Abrol et al., 2020~\cite{abrolDeepResidualLearning2020} & ADNI & T1w & AD classification & \makecell{FM visualization \\ Perturbation} & \ref{sec:application_FM}, \ref{sec:application_peturbation}\\
    \hline
    Bae et al., 2019~\cite{baeTransferLearningPredicting2019} & ADNI & sMRI & AD classification & Perturbation & \ref{sec:application_peturbation} \\
    \hline
    Ball et al., 2020~\cite{ballIndividualVariationUnderlying2020} & PING & T1w & Age prediction & \makecell{Weight visualization \\ SHAP} & \ref{sec:application_weights}, \ref{sec:application_distillation} \\
    \hline
    Biffi et al., 2020~\cite{biffiExplainableAnatomicalShape2020} & ADNI & T1w & AD classification & FM visualization & \ref{sec:application_FM} \\
    \hline
    B\"{o}hle et al., 2019~\cite{bohleLayerwiseRelevancePropagation2019} & ADNI & T1w & AD classification & \makecell{LRP \\ Guided back-propagation} & \ref{sec:application_BP} \\
    \hline
    Burduja et al., 2020~\cite{burdujaAccurateEfficientIntracranial2020} & RSNA & CT scan & Intracranial Hemorrhage detection & Grad-CAM & \ref{sec:application_BP} \\
    \hline
    Cecotti and Gr\"{a}ser, 2011~\cite{cecottiConvolutionalNeuralNetworks2011} & in-house & EEG & P300 signals detection & Weight visualization & \ref{sec:application_weights} \\
    \hline
    Dyrba et al., 2020~\cite{dyrbaComparisonCNNVisualization2020} & ADNI & T1w & AD classification & \makecell{DeconvNet \\ Deep Taylor decomposition \\ Gradient $\odot$ Input \\ LRP \\ Grad-CAM} & \ref{sec:application_BP} \\
    \hline
    Eitel and Ritter, 2019~\cite{eitelTestingRobustnessAttribution2019} & ADNI & T1w & AD classification & \makecell{Gradient $\odot$ Input \\ Guided back-propagation \\ LRP \\ Perturbation} & \ref{sec:application_BP}, \ref{sec:application_peturbation} \\
    \hline
    Eitel et al., 2019~\cite{eitelUncoveringConvolutionalNeural2019} & ADNI,  in-house & T1w & Multiple Sclerosis detection & \makecell{Gradient $\odot$ Input \\ LRP} & \ref{sec:application_BP} \\
    \hline
    
    \end{tabular}
\end{table}

\end{landscape}
\clearpage
}

\afterpage{%
\clearpage
\thispagestyle{empty}
\begin{landscape}

\begin{table}[!t]
        \centering
    \begin{tabular}{|m{0.35\textwidth}<{\centering}|m{0.25\textwidth}<{\centering}|m{0.15\textwidth}<{\centering}|m{0.3\textwidth}<{\centering}|m{0.4\textwidth}<{\centering}|m{0.1\textwidth}<{\centering}|}
    \hline
    Study & Data set & Modality & Task & Interpretability method & Section \\
    \hline
    \hline
    Fu et al., 2021~\cite{fuAttentionbasedFullSlice2021} & CQ500, RSNA & CT scan & Detection of Critical Findings in Head CT scan & Attention mechanism & \ref{sec:application_intrinsic} \\
    \hline
    Guti\'{e}rrez-Becker and Wachinger, 2018~\cite{gutierrez-beckerDeepMultistructuralShape2018} & ADNI & T1w & AD classification & Perturbation & \ref{sec:application_peturbation} \\
    \hline
    Hu et al., 2021~ \cite{huDeepLearningBasedClassification2021} & ADNI, NIFD & T1w & AD/CN/FTD classification & Guided back-propagation & \ref{sec:application_BP} \\
    \hline
    Jin et al., 2020~\cite{jinGeneralizableReproducibleNeuroscientifically2020} & ADNI, in-house & T1w & AD classification & Attention mechanism & \ref{sec:application_intrinsic} \\
    \hline
    Lee et al., 2019~\cite{leeInterpretableAlzheimerDisease2019} & ADNI & T1w & AD classification & Modular transparency & \ref{sec:application_intrinsic} \\
    \hline
    Leming et al., 2020~\cite{lemingEnsembleDeepLearning2020} & OpenFMRI, ADNI, ABIDE, ABIDE II, ABCD, NDAR ICBM, UK Biobank, 1000FC & fMRI & Autism~classification Sex~classification Task~vs~rest classification & \makecell{FM visualization \\ Grad-CAM} & \ref{sec:application_FM}, \ref{sec:application_BP} \\
    \hline
    Magesh et al., 2020~\cite{mageshExplainableMachineLearning2020} & PPMI & SPECT & Parkinson's disease detection & LIME & \ref{sec:application_distillation} \\
    \hline
    Martinez-Murcia et al., 2020~\cite{martinez-murciaStudyingManifoldStructure2020} & ADNI & T1w & AD~classification Prediction~of neuropsychological tests \& other clinical variables & FM visualization & \ref{sec:application_FM} \\
    \hline
    Nigri et al., 2020~\cite{nigriExplainableDeepCNNs2020} & ADNI, AIBL & T1w & AD classification & \makecell{Perturbation \\ Swap test} & \ref{sec:application_peturbation} \\
    \hline
    Oh et al., 2019~\cite{ohClassificationVisualizationAlzheimer2019} & ADNI & T1w & AD classification & \makecell{FM visualization \\ Standard back-propagation \\ Perturbation} & \ref{sec:application_FM}, \ref{sec:application_BP}, \ref{sec:application_peturbation} \\
    \hline
    
    \end{tabular}
\end{table}

\end{landscape}
\clearpage
}

\afterpage{%
\clearpage
\thispagestyle{empty}
\begin{landscape}

\begin{table}[!t]
    {\centering
    \begin{tabular}{|m{0.35\textwidth}<{\centering}|m{0.25\textwidth}<{\centering}|m{0.15\textwidth}<{\centering}|m{0.3\textwidth}<{\centering}|m{0.4\textwidth}<{\centering}|m{0.1\textwidth}<{\centering}|}
    \hline
    Study & Data set & Modality & Task & Interpretability method & Section \\
    \hline
    \hline
    Qiu et al., 2020~\cite{qiuDevelopmentValidationInterpretable2020} & ADNI, AIBL, FHS, NACC & T1w & AD classification & Modular transparency & \ref{sec:application_intrinsic} \\
    \hline
    Ravi et al., 2022~\cite{raviDegenerativeAdversarialNeuroimage2022} & ADNI & T1w & CN/MCI/AD reconstruction & Modular transparency & \ref{sec:application_intrinsic} \\
    \hline
    Rieke et al., 2018~\cite{riekeVisualizingConvolutionalNetworks2018} & ADNI & T1w & AD classification & \makecell{Standard back-propagation \\ Guided back-propagation \\ Perturbation \\ Brain area occlusion} & \ref{sec:application_BP}, \ref{sec:application_peturbation} \\
    \hline
    Tang et al., 2019~\cite{tangInterpretableClassificationAlzheimer2019} & UCD-ADC, Brain Bank & Histology & Detection of  amyloid-$\beta$ pathology & \makecell{Guided back-propagation \\ Perturbation} & \ref{sec:application_BP}, \ref{sec:application_peturbation} \\
    \hline
    Wood et al., 2019~\cite{woodNEURODRAM3DRecurrent2019} & ADNI & T1w & AD classification & Modular transparency & \ref{sec:application_intrinsic} \\
    \hline
    \end{tabular}
    }

    \bigskip
    \textbf{Data sets}: 1000FC, 1000 Functional Connectomes; ABCD, Adolescent Brain Cognitive Development; ABIDE, Autism Brain Imaging Data Exchange; ADNI, Alzheimer’s Disease Neuroimaging Initiative; AIBL, Australian Imaging, Biomarkers and Lifestyle; FHS, Framingham Heart Study; ICBM, International Consortium for Brain Mapping; NACC, National Alzheimer’s Coordinating Center; NDAR, National Database for Autism Research; NIFD, frontotemporal lobar degeneration neuroimaging initiative; PING, Pediatric Imaging, Neurocognition and Genetics; PPMI, Parkinson’s Progression Markers Initiative; RSNA, Radiological Society of North America 2019 Brain CT Hemorrhage dataset; UCD-ADC Brain Bank, University of California Davis Alzheimer’s Disease Center Brain Bank. \\
    \textbf{Modalities}: CT, computed tomography; EEG, electroencephalography; fMRI, functional magnetic resonance imaging; sMRI, structural magnetic resonance imaging; SPECT,
    single-photon emission computed tomography; T1w, T1-weighted [magnetic resonance imaging]. \\
    \textbf{Tasks}: AD, Alzheimer's disease; CN, cognitively normal; FTD, fronto-temporal dementia; MCI, mild cognitive impairment. \\
    \textbf{Interpretability methods}: FM, feature maps; Grad-CAM, gradient-weighted class activation mapping; LIME, local interpretable model-agnostic explanations; LRP, layer-wise relevance; SHAP, SHapley Additive exPlanations.
\end{table}

\end{landscape}
\clearpage
}

%% file: chapter22-Thibeau-Sutre_Collin-interpretability.bbl
\begin{thebibliography}{54}
\providecommand{\natexlab}[1]{#1}
\providecommand{\url}[1]{{#1}}
\providecommand{\urlprefix}{URL }
\expandafter\ifx\csname urlstyle\endcsname\relax
  \providecommand{\doi}[1]{DOI~\discretionary{}{}{}#1}\else
  \providecommand{\doi}{DOI~\discretionary{}{}{}\begingroup
  \urlstyle{rm}\Url}\fi
\providecommand{\eprint}[2][]{\url{#2}}

\bibitem[{Ribeiro et~al(2016)Ribeiro, Singh, and
  Guestrin}]{ribeiroWhyShouldTrust2016}
Ribeiro MT, Singh S, Guestrin C (2016) "{{Why Should I Trust You}}?":
  {{Explaining}} the {{Predictions}} of {{Any Classifier}}. In: Proceedings of
  the 22nd {{ACM SIGKDD International Conference}} on {{Knowledge Discovery}}
  and {{Data Mining}} - {{KDD}} '16, {ACM Press}, {San Francisco, California,
  USA}, pp 1135--1144, \doi{10.1145/2939672.2939778}

\bibitem[{Fong and Vedaldi(2017)}]{fongInterpretableExplanationsBlack2017}
Fong RC, Vedaldi A (2017) Interpretable {{Explanations}} of {{Black Boxes}} by
  {{Meaningful Perturbation}}. In: 2017 {{IEEE International Conference}} on
  {{Computer Vision}} ({{ICCV}}), pp 3449--3457, \doi{10.1109/ICCV.2017.371}

\bibitem[{DeGrave et~al(2021)DeGrave, Janizek, and
  Lee}]{degraveAIRadiographicCOVID192021}
DeGrave AJ, Janizek JD, Lee SI (2021) {{AI}} for radiographic {{COVID}}-19
  detection selects shortcuts over signal. Nature Machine Intelligence
  3(7):610--619, \doi{10.1038/s42256-021-00338-7}

\bibitem[{Lipton(2018)}]{liptonMythosModelInterpretability2018}
Lipton ZC (2018) The mythos of model interpretability. Communications of the
  ACM 61(10):36--43, \doi{10.1145/3233231}

\bibitem[{Xie et~al(2020)Xie, Ras, {van Gerven}, and
  Doran}]{xieExplainableDeepLearning2020}
Xie N, Ras G, {van Gerven} M, Doran D (2020) Explainable {{Deep Learning}}: {{A
  Field Guide}} for the {{Uninitiated}}. arXiv:200414545 [cs, stat]
  \eprint{2004.14545}

\bibitem[{Adebayo et~al(2018)Adebayo, Gilmer, Muelly, Goodfellow, Hardt, and
  Kim}]{adebayoSanityChecksSaliency2018}
Adebayo J, Gilmer J, Muelly M, Goodfellow I, Hardt M, Kim B (2018) Sanity
  checks for saliency maps. In: Advances in {{Neural Information Processing
  Systems}}, pp 9505--9515

\bibitem[{Krizhevsky et~al(2012)Krizhevsky, Sutskever, and
  Hinton}]{krizhevskyImageNetClassificationDeep2012}
Krizhevsky A, Sutskever I, Hinton GE (2012) {{ImageNet Classification}} with
  {{Deep Convolutional Neural Networks}}. In: Pereira F, Burges CJC, Bottou L,
  Weinberger KQ (eds) Advances in {{Neural Information Processing Systems}} 25,
  {Curran Associates, Inc.}, pp 1097--1105

\bibitem[{Voss et~al(2021)Voss, Cammarata, Goh, Petrov, Schubert, Egan, Lim,
  and Olah}]{vossVisualizingWeights2021}
Voss C, Cammarata N, Goh G, Petrov M, Schubert L, Egan B, Lim SK, Olah C (2021)
  Visualizing {{Weights}}. Distill 6(2):e00024.007,
  \doi{10.23915/distill.00024.007}

\bibitem[{Olah et~al(2017)Olah, Mordvintsev, and
  Schubert}]{olahFeatureVisualization2017a}
Olah C, Mordvintsev A, Schubert L (2017) Feature {{Visualization}}. Distill
  2(11):e7, \doi{10.23915/distill.00007}

\bibitem[{Simonyan et~al(2013)Simonyan, Vedaldi, and
  Zisserman}]{simonyanDeepConvolutionalNetworks2013}
Simonyan K, Vedaldi A, Zisserman A (2013) Deep {{Inside Convolutional
  Networks}}: {{Visualising Image Classification Models}} and {{Saliency
  Maps}}. arXiv:13126034 [cs] \eprint{1312.6034}

\bibitem[{Shrikumar et~al(2017)Shrikumar, Greenside, Shcherbina, and
  Kundaje}]{shrikumarNotJustBlack2017a}
Shrikumar A, Greenside P, Shcherbina A, Kundaje A (2017) Not {{Just}} a {{Black
  Box}}: {{Learning Important Features Through Propagating Activation
  Differences}}. arXiv:160501713 [cs] \eprint{1605.01713}

\bibitem[{Springenberg et~al(2014)Springenberg, Dosovitskiy, Brox, and
  Riedmiller}]{springenbergStrivingSimplicityAll2014}
Springenberg JT, Dosovitskiy A, Brox T, Riedmiller M (2014) Striving for
  {{Simplicity}}: {{The All Convolutional Net}}. arXiv:14126806 [cs]
  \eprint{1412.6806}

\bibitem[{Zhou et~al(2015)Zhou, Khosla, Lapedriza, Oliva, and
  Torralba}]{zhouLearningDeepFeatures2015}
Zhou B, Khosla A, Lapedriza A, Oliva A, Torralba A (2015) Learning {{Deep
  Features}} for {{Discriminative Localization}}. arXiv:151204150 [cs]
  \eprint{1512.04150}

\bibitem[{Selvaraju et~al(2017)Selvaraju, Cogswell, Das, Vedantam, Parikh, and
  Batra}]{selvarajuGradCAMVisualExplanations2017}
Selvaraju RR, Cogswell M, Das A, Vedantam R, Parikh D, Batra D (2017)
  Grad-{{CAM}}: {{Visual Explanations}} from {{Deep Networks}} via
  {{Gradient}}-{{Based Localization}}. In: 2017 {{IEEE International
  Conference}} on {{Computer Vision}} ({{ICCV}}), pp 618--626,
  \doi{10.1109/ICCV.2017.74}

\bibitem[{Bach et~al(2015)Bach, Binder, Montavon, Klauschen, M{\"u}ller, and
  Samek}]{bachPixelWiseExplanationsNonLinear2015}
Bach S, Binder A, Montavon G, Klauschen F, M{\"u}ller KR, Samek W (2015) On
  {{Pixel}}-{{Wise Explanations}} for {{Non}}-{{Linear Classifier Decisions}}
  by {{Layer}}-{{Wise Relevance Propagation}}. PLOS ONE 10(7):e0130140,
  \doi{10.1371/journal.pone.0130140}

\bibitem[{Samek et~al(2017)Samek, Binder, Montavon, Lapuschkin, and
  M{\"u}ller}]{samekEvaluatingVisualizationWhat2017}
Samek W, Binder A, Montavon G, Lapuschkin S, M{\"u}ller KR (2017) Evaluating
  the {{Visualization}} of {{What}} a {{Deep Neural Network Has Learned}}. IEEE
  Transactions on Neural Networks and Learning Systems 28(11):2660--2673,
  \doi{10.1109/TNNLS.2016.2599820}

\bibitem[{Montavon et~al(2017)Montavon, Lapuschkin, Binder, Samek, and
  M{\"u}ller}]{montavonExplainingNonlinearClassification2017}
Montavon G, Lapuschkin S, Binder A, Samek W, M{\"u}ller KR (2017) Explaining
  nonlinear classification decisions with deep {{Taylor}} decomposition.
  Pattern Recognition 65:211--222, \doi{10.1016/j.patcog.2016.11.008}

\bibitem[{Montavon et~al(2018)Montavon, Samek, and
  M{\"u}ller}]{montavonMethodsInterpretingUnderstanding2018}
Montavon G, Samek W, M{\"u}ller KR (2018) Methods for interpreting and
  understanding deep neural networks. Digital Signal Processing 73:1--15,
  \doi{10.1016/j.dsp.2017.10.011}

\bibitem[{Zeiler and
  Fergus(2014)}]{zeilerVisualizingUnderstandingConvolutional2014}
Zeiler MD, Fergus R (2014) Visualizing and {{Understanding Convolutional
  Networks}}. In: Fleet D, Pajdla T, Schiele B, Tuytelaars T (eds) Computer
  {{Vision}} \textendash{} {{ECCV}} 2014, {Springer International Publishing},
  Lecture {{Notes}} in {{Computer Science}}, pp 818--833

\bibitem[{Lundberg and Lee(2017)}]{lundbergUnifiedApproachInterpreting2017a}
Lundberg SM, Lee SI (2017) {A Unified Approach to Interpreting Model
  Predictions}. In: Proceedings of the 31st {{International Conference}} on
  {{Neural Information Processing Systems}}, {Curran Associates Inc.}, {Red
  Hook, NY, USA}, {{NIPS}}'17, pp 4768--4777

\bibitem[{Frosst and Hinton(2017)}]{frosstDistillingNeuralNetwork2017}
Frosst N, Hinton G (2017) Distilling a {{Neural Network Into}} a {{Soft
  Decision Tree}}. arXiv:171109784 [cs, stat] \eprint{1711.09784}

\bibitem[{Xu et~al(2016)Xu, Ba, Kiros, Cho, Courville, Salakhutdinov, Zemel,
  and Bengio}]{xuShowAttendTell2016}
Xu K, Ba J, Kiros R, Cho K, Courville A, Salakhutdinov R, Zemel R, Bengio Y
  (2016) Show, {{Attend}} and {{Tell}}: {{Neural Image Caption Generation}}
  with {{Visual Attention}}. arXiv:150203044 [cs] \eprint{1502.03044}

\bibitem[{Wang et~al(2017)Wang, Jiang, Qian, Yang, Li, Zhang, Wang, and
  Tang}]{wangResidualAttentionNetwork2017a}
Wang F, Jiang M, Qian C, Yang S, Li C, Zhang H, Wang X, Tang X (2017) Residual
  {{Attention Network}} for {{Image Classification}}. In: 2017 {{IEEE
  Conference}} on {{Computer Vision}} and {{Pattern Recognition}} ({{CVPR}}),
  {IEEE}, {Honolulu, HI}, pp 6450--6458, \doi{10.1109/CVPR.2017.683}

\bibitem[{Ba et~al(2015)Ba, Mnih, and
  Kavukcuoglu}]{baMultipleObjectRecognition2015}
Ba J, Mnih V, Kavukcuoglu K (2015) Multiple {{Object Recognition}} with
  {{Visual Attention}}. arXiv:14127755 [cs] \eprint{1412.7755}

\bibitem[{Yeh et~al(2019)Yeh, Hsieh, Suggala, Inouye, and
  Ravikumar}]{yehFidelitySensitivityExplanations2019}
Yeh CK, Hsieh CY, Suggala A, Inouye DI, Ravikumar PK (2019) On the
  ({{In}})fidelity and {{Sensitivity}} of {{Explanations}}. In: Wallach H,
  Larochelle H, Beygelzimer A, d{\textbackslash}textquotesingle {Alch{\'e}-Buc}
  F, Fox E, Garnett R (eds) Advances in {{Neural Information Processing
  Systems}} 32, {Curran Associates, Inc.}, pp 10967--10978

\bibitem[{Cecotti and
  Gr{\"a}ser(2011)}]{cecottiConvolutionalNeuralNetworks2011}
Cecotti H, Gr{\"a}ser A (2011) Convolutional neural networks for {{P300}}
  detection with application to brain-computer interfaces. IEEE Transactions on
  Pattern Analysis and Machine Intelligence 33(3):433--445,
  \doi{10.1109/TPAMI.2010.125}

\bibitem[{Oh et~al(2019)Oh, Chung, Kim, Kim, and
  Oh}]{ohClassificationVisualizationAlzheimer2019}
Oh K, Chung YC, Kim KW, Kim WS, Oh IS (2019) Classification and
  {{Visualization}} of {{Alzheimer}}’s {{Disease}} using {{Volumetric
  Convolutional Neural Network}} and {{Transfer Learning}}. Scientific Reports
  9(1):1--16, \doi{10.1038/s41598-019-54548-6}

\bibitem[{Abrol et~al(2020)Abrol, Bhattarai, Fedorov, Du, Plis, and
  Calhoun}]{abrolDeepResidualLearning2020}
Abrol A, Bhattarai M, Fedorov A, Du Y, Plis S, Calhoun V (2020) Deep residual
  learning for neuroimaging: {{An}} application to predict progression to
  {{Alzheimer}}’s disease. Journal of Neuroscience Methods 339:108701,
  \doi{10.1016/j.jneumeth.2020.108701}

\bibitem[{Biffi et~al(2020)Biffi, Cerrolaza, Tarroni, Bai, De~Marvao, Oktay,
  Ledig, Le~Folgoc, Kamnitsas, Doumou, Duan, Prasad, Cook, O'Regan, and
  Rueckert}]{biffiExplainableAnatomicalShape2020}
Biffi C, Cerrolaza J, Tarroni G, Bai W, De~Marvao A, Oktay O, Ledig C,
  Le~Folgoc L, Kamnitsas K, Doumou G, Duan J, Prasad S, Cook S, O'Regan D,
  Rueckert D (2020) Explainable {{Anatomical Shape Analysis}} through {{Deep
  Hierarchical Generative Models}}. IEEE Transactions on Medical Imaging
  39(6):2088--2099, \doi{10.1109/TMI.2020.2964499}

\bibitem[{{Martinez-Murcia} et~al(2020){Martinez-Murcia}, Ortiz, Gorriz,
  Ramirez, and
  {Castillo-Barnes}}]{martinez-murciaStudyingManifoldStructure2020}
{Martinez-Murcia} FJ, Ortiz A, Gorriz JM, Ramirez J, {Castillo-Barnes} D (2020)
  Studying the {{Manifold Structure}} of {{Alzheimer}}'s {{Disease}}: {{A Deep
  Learning Approach Using Convolutional Autoencoders}}. IEEE Journal of
  Biomedical and Health Informatics 24(1):17--26,
  \doi{10.1109/JBHI.2019.2914970}

\bibitem[{Leming et~al(2020)Leming, Górriz, and
  Suckling}]{lemingEnsembleDeepLearning2020}
Leming M, Górriz JM, Suckling J (2020) Ensemble {{Deep Learning}} on
  {{Large}}, {{Mixed}}-{{Site fMRI Datasets}} in {{Autism}} and {{Other
  Tasks}}. International Journal of Neural Systems p 2050012,
  \doi{10.1142/S0129065720500124}, \eprint{2002.07874}

\bibitem[{Bae et~al(2019)Bae, Stocks, Heywood, Jung, Jenkins, Katsaggelos,
  Popuri, Beg, and Wang}]{baeTransferLearningPredicting2019}
Bae J, Stocks J, Heywood A, Jung Y, Jenkins L, Katsaggelos A, Popuri K, Beg MF,
  Wang L (2019) Transfer {{Learning}} for {{Predicting Conversion}} from {{Mild
  Cognitive Impairment}} to {{Dementia}} of {{Alzheimer}}’s {{Type}} based on
  {{3D}}-{{Convolutional Neural Network}}. bioRxiv
  \doi{10.1101/2019.12.20.884932}

\bibitem[{Ball et~al(2021)Ball, Kelly, Beare, and
  Seal}]{ballIndividualVariationUnderlying2020}
Ball G, Kelly CE, Beare R, Seal ML (2021) Individual variation underlying brain
  age estimates in typical development. NeuroImage 235:118036,
  \doi{10.1016/j.neuroimage.2021.118036}

\bibitem[{Böhle et~al(2019)Böhle, Eitel, Weygandt, Ritter, and
  {on}}]{bohleLayerwiseRelevancePropagation2019}
Böhle M, Eitel F, Weygandt M, Ritter K, {on} botADNI (2019) Layer-wise
  relevance propagation for explaining deep neural network decisions in
  {{MRI}}-based {{Alzheimer}}'s disease classification. Frontiers in Aging
  Neuroscience 10(JUL), \doi{10.3389/fnagi.2019.00194}

\bibitem[{Burduja et~al(2020)Burduja, Ionescu, and
  Verga}]{burdujaAccurateEfficientIntracranial2020}
Burduja M, Ionescu RT, Verga N (2020) Accurate and {{Efficient Intracranial
  Hemorrhage Detection}} and {{Subtype Classification}} in {{3D CT Scans}} with
  {{Convolutional}} and {{Long Short}}-{{Term Memory Neural Networks}}. Sensors
  20(19):5611, \doi{10.3390/s20195611}

\bibitem[{Dyrba et~al(2020)Dyrba, Pallath, and
  Marzban}]{dyrbaComparisonCNNVisualization2020}
Dyrba M, Pallath AH, Marzban EN (2020) {Comparison of CNN Visualization Methods
  to Aid Model Interpretability for Detecting Alzheimer’s Disease}. In:
  Tolxdorff T, Deserno TM, Handels H, Maier A, {Maier-Hein} KH, Palm C (eds)
  {Bildverarbeitung für die Medizin 2020}, {Springer Fachmedien}, {Wiesbaden},
  {Informatik aktuell}, pp 307--312, \doi{10.1007/978-3-658-29267-6_68}

\bibitem[{Eitel and Ritter(2019)}]{eitelTestingRobustnessAttribution2019}
Eitel F, Ritter K (2019) Testing the {{Robustness}} of {{Attribution Methods}}
  for {{Convolutional Neural Networks}} in {{MRI}}-{{Based Alzheimer}}’s
  {{Disease Classification}}. In: Interpretability of {{Machine Intelligence}}
  in {{Medical Image Computing}} and {{Multimodal Learning}} for {{Clinical
  Decision Support}}, {Springer International Publishing}, {Cham}, Lecture
  {{Notes}} in {{Computer Science}}, pp 3--11,
  \doi{10.1007/978-3-030-33850-3_1}

\bibitem[{Eitel et~al(2019)Eitel, Soehler, {Bellmann-Strobl}, Brandt, Ruprecht,
  Giess, Kuchling, Asseyer, Weygandt, Haynes, Scheel, Paul, and
  Ritter}]{eitelUncoveringConvolutionalNeural2019}
Eitel F, Soehler E, {Bellmann-Strobl} J, Brandt AU, Ruprecht K, Giess RM,
  Kuchling J, Asseyer S, Weygandt M, Haynes JD, Scheel M, Paul F, Ritter K
  (2019) Uncovering convolutional neural network decisions for diagnosing
  multiple sclerosis on conventional {{MRI}} using layer-wise relevance
  propagation. NeuroImage: Clinical 24:102003, \doi{10.1016/j.nicl.2019.102003}

\bibitem[{Fu et~al(2021)Fu, Li, Wang, Ma, and
  Chen}]{fuAttentionbasedFullSlice2021}
Fu G, Li J, Wang R, Ma Y, Chen Y (2021) Attention-based full slice brain {{CT}}
  image diagnosis with explanations. Neurocomputing 452:263--274,
  \doi{10.1016/j.neucom.2021.04.044}

\bibitem[{{Gutiérrez-Becker} and
  Wachinger(2018)}]{gutierrez-beckerDeepMultistructuralShape2018}
{Gutiérrez-Becker} B, Wachinger C (2018) Deep multi-structural shape analysis:
  {{Application}} to neuroanatomy. In: Lecture {{Notes}} in {{Computer
  Science}} (Including Subseries {{Lecture Notes}} in {{Artificial
  Intelligence}} and {{Lecture Notes}} in {{Bioinformatics}}), vol 11072 LNCS,
  pp 523--531, \doi{10.1007/978-3-030-00931-1_60}

\bibitem[{Hu et~al(2021)Hu, Qing, Liu, Zhang, Lv, Wang, Wang, He, Gao, and
  Zhang}]{huDeepLearningBasedClassification2021}
Hu J, Qing Z, Liu R, Zhang X, Lv P, Wang M, Wang Y, He K, Gao Y, Zhang B (2021)
  Deep {{Learning}}-{{Based Classification}} and {{Voxel}}-{{Based
  Visualization}} of {{Frontotemporal Dementia}} and {{Alzheimer}}’s
  {{Disease}}. Frontiers in Neuroscience 14, \doi{10.3389/fnins.2020.626154}

\bibitem[{Jin et~al(2020)Jin, Zhou, Han, Ren, Han, Liu, Lu, Song, Wang, Wang,
  Xu, Yang, Yao, Yu, Zhao, Wintermark, Zuo, Zhang, Zhou, Zhang, Jiang, Wang,
  and Liu}]{jinGeneralizableReproducibleNeuroscientifically2020}
Jin D, Zhou B, Han Y, Ren J, Han T, Liu B, Lu J, Song C, Wang P, Wang D, Xu J,
  Yang Z, Yao H, Yu C, Zhao K, Wintermark M, Zuo N, Zhang X, Zhou Y, Zhang X,
  Jiang T, Wang Q, Liu Y (2020) Generalizable, {{Reproducible}}, and
  {{Neuroscientifically Interpretable Imaging Biomarkers}} for {{Alzheimer}}'s
  {{Disease}}. Advanced Science 7(14):2000675, \doi{10.1002/advs.202000675}

\bibitem[{Lee et~al(2019)Lee, Choi, Kim, Suk, and {Alzheimer’s Disease
  Neuroimaging Initiative}}]{leeInterpretableAlzheimerDisease2019}
Lee E, Choi JS, Kim M, Suk HI, {Alzheimer’s Disease Neuroimaging Initiative}
  (2019) Toward an interpretable {{Alzheimer}}'s disease diagnostic model with
  regional abnormality representation via deep learning. NeuroImage 202:116113,
  \doi{10.1016/j.neuroimage.2019.116113}

\bibitem[{Magesh et~al(2020)Magesh, Myloth, and
  Tom}]{mageshExplainableMachineLearning2020}
Magesh PR, Myloth RD, Tom RJ (2020) An {{Explainable Machine Learning Model}}
  for {{Early Detection}} of {{Parkinson}}'s {{Disease}} using {{LIME}} on
  {{DaTSCAN Imagery}}. Computers in Biology and Medicine 126:104041,
  \doi{10.1016/j.compbiomed.2020.104041}

\bibitem[{Nigri et~al(2020)Nigri, Ziviani, Cappabianco, Antunes, and
  Veloso}]{nigriExplainableDeepCNNs2020}
Nigri E, Ziviani N, Cappabianco F, Antunes A, Veloso A (2020) Explainable
  {{Deep CNNs}} for {{MRI}}-{{Based Diagnosis}} of {{Alzheimer}}'s {{Disease}}.
  In: Proceedings of the {{International Joint Conference}} on {{Neural
  Networks}}, \doi{10.1109/IJCNN48605.2020.9206837}

\bibitem[{Qiu et~al(2020)Qiu, Joshi, Miller, Xue, Zhou, Karjadi, Chang, Joshi,
  Dwyer, Zhu, Kaku, Zhou, Alderazi, Swaminathan, Kedar, {Saint-Hilaire},
  Auerbach, Yuan, Sartor, Au, and
  Kolachalama}]{qiuDevelopmentValidationInterpretable2020}
Qiu S, Joshi PS, Miller MI, Xue C, Zhou X, Karjadi C, Chang GH, Joshi AS, Dwyer
  B, Zhu S, Kaku M, Zhou Y, Alderazi YJ, Swaminathan A, Kedar S,
  {Saint-Hilaire} MH, Auerbach SH, Yuan J, Sartor EA, Au R, Kolachalama VB
  (2020) Development and validation of an interpretable deep learning framework
  for {{Alzheimer}}'s disease classification. Brain: A Journal of Neurology
  143(6):1920--1933, \doi{10.1093/brain/awaa137}

\bibitem[{Ravi et~al(2022)Ravi, Blumberg, Ingala, Barkhof, Alexander, and
  Oxtoby}]{raviDegenerativeAdversarialNeuroimage2022}
Ravi D, Blumberg SB, Ingala S, Barkhof F, Alexander DC, Oxtoby NP (2022)
  Degenerative adversarial neuroimage nets for brain scan simulations:
  Application in ageing and dementia. Medical Image Analysis 75:102257,
  \doi{10.1016/j.media.2021.102257}

\bibitem[{Rieke et~al(2018)Rieke, Eitel, Weygandt, Haynes, and
  Ritter}]{riekeVisualizingConvolutionalNetworks2018}
Rieke J, Eitel F, Weygandt M, Haynes JD, Ritter K (2018) Visualizing
  {{Convolutional Networks}} for {{MRI}}-{{Based Diagnosis}} of
  {{Alzheimer}}’s {{Disease}}. In: Understanding and {{Interpreting Machine
  Learning}} in {{Medical Image Computing Applications}}, {Springer
  International Publishing}, {Cham}, Lecture {{Notes}} in {{Computer Science}},
  pp 24--31, \doi{10.1007/978-3-030-02628-8_3}

\bibitem[{Tang et~al(2019)Tang, Chuang, DeCarli, Jin, Beckett, Keiser, and
  Dugger}]{tangInterpretableClassificationAlzheimer2019}
Tang Z, Chuang KV, DeCarli C, Jin LW, Beckett L, Keiser MJ, Dugger BN (2019)
  Interpretable classification of {{Alzheimer}}’s disease pathologies with a
  convolutional neural network pipeline. Nature Communications 10(1):1--14,
  \doi{10.1038/s41467-019-10212-1}

\bibitem[{Wood et~al(2019)Wood, Cole, and Booth}]{woodNEURODRAM3DRecurrent2019}
Wood D, Cole J, Booth T (2019) {{NEURO}}-{{DRAM}}: A {{3D}} recurrent visual
  attention model for interpretable neuroimaging classification.
  arXiv:191004721 [cs, stat] \eprint{1910.04721}

\bibitem[{{Thibeau-Sutre} et~al(2020){Thibeau-Sutre}, Colliot, Dormont, and
  Burgos}]{thibeau-sutreVisualizationApproachAssess2020}
{Thibeau-Sutre} E, Colliot O, Dormont D, Burgos N (2020) Visualization approach
  to assess the robustness of neural networks for medical image classification.
  In: Medical {{Imaging}} 2020: {{Image Processing}}, {International Society
  for Optics and Photonics}, vol 11313, p 113131J, \doi{10.1117/12.2548952}

\bibitem[{Tomsett et~al(2020)Tomsett, Harborne, Chakraborty, Gurram, and
  Preece}]{tomsettSanityChecksSaliency2020}
Tomsett R, Harborne D, Chakraborty S, Gurram P, Preece A (2020) Sanity
  {{Checks}} for {{Saliency Metrics}}. Proceedings of the AAAI Conference on
  Artificial Intelligence 34(04):6021--6029, \doi{10.1609/aaai.v34i04.6064}

\bibitem[{Lapuschkin et~al(2016)Lapuschkin, Binder, Montavon, Muller, and
  Samek}]{lapuschkinAnalyzingClassifiersFisher2016}
Lapuschkin S, Binder A, Montavon G, Muller KR, Samek W (2016) Analyzing
  {{Classifiers}}: {{Fisher Vectors}} and {{Deep Neural Networks}}. In: 2016
  {{IEEE Conference}} on {{Computer Vision}} and {{Pattern Recognition}}
  ({{CVPR}}), {IEEE}, {Las Vegas, NV, USA}, pp 2912--2920,
  \doi{10.1109/CVPR.2016.318}

\bibitem[{Ozbulak(2019)}]{uozbulak_pytorch_vis_2021}
Ozbulak U (2019) Pytorch cnn visualizations.
  \url{https://github.com/utkuozbulak/pytorch-cnn-visualizations}

\end{thebibliography}
